\documentclass[twoside,11pt]{article}
\usepackage[preprint]{jmlr2e_mod}

\usepackage{microtype}
\usepackage{graphicx}
\usepackage{subfigure}
\usepackage{booktabs} 

\usepackage{amsmath, amsfonts}
\usepackage{url}
\usepackage{color}
\usepackage[usenames,dvipsnames,svgnames,table]{xcolor}
\usepackage{booktabs}
\usepackage{bm}
\usepackage{stackrel}
\usepackage{enumitem}

\usepackage{algorithm}
\usepackage[noend]{algpseudocode}

\usepackage[export]{adjustbox}

\newcommand{\Rbb}{\mathbb{R}}

\newcommand{\Acal}{\mathcal{A}}

\newcommand{\Dcal}{\mathcal{D}}

\newcommand{\Ncal}{\mathcal{N}}
\newcommand{\Ocal}{\mathcal{O}}

\newcommand{\Xcal}{\mathcal{X}}
\newcommand{\Ycal}{\mathcal{Y}}

\def\expec#1#2{{\mathbb{E}}_{#1}[ #2 ]}
\def\expecf#1#2{{\mathbb{E}}_{#1}\left[ #2 \right]}

\def\entropy#1{\mbox{H}[ #1 ]}
\def\entropyf#1{\mbox{H}\left[ #1 \right]}

\newcommand{\infobax}{\textsc{InfoBAX }}
\newcommand{\infobaxnospace}{\textsc{InfoBAX}}
\newcommand{\randomsearch}{\textsc{RandomSearch }}
\newcommand{\randomsearchnospace}{\textsc{RandomSearch}}
\newcommand{\uncertaintysampling}{\textsc{UncertaintySampling }}
\newcommand{\uncertaintysamplingnospace}{\textsc{UncertaintySampling}}
\newcommand{\dataset}{\Dcal}
\newcommand{\datasett}{\dataset_t}

\DeclareMathOperator{\EIG}{EIG}

\newcommand{\acqfn}{\EIG}
\newcommand{\acqfnt}{\acqfn_t}

\def\samp#1{\widetilde{#1}}

\newcommand{\outputplain}{O}
\newcommand{\algoutput}{\outputplain_\Acal}

\newcommand{\algoutputsamp}{\samp{\outputplain}_\Acal}

\newcommand{\algoutputring}{\mathring{\outputplain}_\Acal}
\newcommand{\outputplainsamp}{\samp{\outputplain}}

\newcommand{\exepathplain}{e}
\newcommand{\exepath}{\exepathplain_\Acal}

\newcommand{\exepathsamp}{\samp{\exepathplain}_\Acal}

\newcommand{\exepathring}{\mathring{\exepathplain}_\Acal}

\newcommand{\subexepathplain}{v}
\newcommand{\subexepath}{\subexepathplain_\Acal}

\newcommand{\subexepathsamp}{\samp{\subexepathplain}_\Acal}
\newcommand{\subexepathplainsamp}{\samp{\subexepathplain}}

\newcommand{\subseqexepathplain}{s}
\newcommand{\subseqexepath}{\subseqexepathplain_\Acal}

\newcommand\given[1][]{\:#1\vert\:}
\newcommand*\diff{\mathop{}\!\mathrm{d}}

\newcommand{\bigCI}{\mathrel{\text{\scalebox{1.07}{$\perp\mkern-10mu\perp$}}}}

\DeclareMathOperator*{\argmax}{arg\,max}

\definecolor{goldenyellow}{rgb}{1.0, 0.87, 0.0}
\definecolor{goldenpoppy}{rgb}{0.99, 0.76, 0.0}

\def\genbox#1#2#3#4#5#6{
    \leavevmode\raise#4bp\hbox to#5bp{\vrule height#5bp depth0bp width0bp
    \pdfliteral{q .5 w \csname #2COLOR\endcsname\space RG
                       \csname #3PDF\endcsname{#5}{#6} S Q
             \ifx1#1 q \csname #2COLOR\endcsname\space rg 
                       \csname #3PDF\endcsname{#5}{#6} f Q\fi}\hss}}

\def\sqbox      #1#2{\genbox{#1}{#2}  {sq}       {0}   {4.5}  {2.25}}

\def\circbox    #1#2{\genbox{#1}{#2}  {circ}     {0}   {5}    {2.5}}

\input{widebar}

\title{Bayesian Algorithm Execution: Estimating Computable Properties of Black-box Functions Using Mutual Information}

\author{\name Willie Neiswanger \email neiswanger@cs.stanford.edu \\
    \addr Stanford University\\
    \AND
    \name Ke Alexander Wang \email alxwang@cs.stanford.edu \\
    \addr Stanford University\\
    \AND
    \name Stefano Ermon \email ermon@cs.stanford.edu \\
    \addr Stanford University\\
}

\ShortHeadings{Bayesian Algorithm Execution}{Neiswanger, Wang, Ermon}

\begin{document}

\maketitle

\vspace{-8mm}
\begin{abstract}
In many real-world problems, we want to infer some property of an expensive black-box function $f$, given a budget of $T$ function evaluations.
One example is budget constrained global optimization of $f$, for which Bayesian optimization is a popular method.
Other properties of interest include local optima, level sets, integrals, or graph-structured information induced by $f$.
Often, we can find an algorithm $\Acal$ to compute the desired property, but it may require far more than $T$ queries to execute.
Given such an $\Acal$, and a prior distribution over $f$, we refer to the problem of inferring the output of $\Acal$ using $T$ evaluations as Bayesian Algorithm Execution (BAX).
To tackle this problem, we present a procedure, \infobaxnospace, that sequentially chooses queries that maximize mutual information with respect to the algorithm's output.
Applying this to Dijkstra’s algorithm, for instance, we infer shortest paths in synthetic and real-world graphs with black-box edge costs.
Using evolution strategies, we yield variants of Bayesian optimization that target local, rather than global, optima.
On these problems, \infobax uses up to 500 times fewer queries to $f$ than required by the original algorithm.
Our method is closely connected to other Bayesian optimal experimental design procedures such as entropy search methods and optimal sensor placement using Gaussian processes.\footnote{See the project website here: \url{https://willieneis.github.io/bax-website}}\footnote{Our code is availabile here: \url{https://github.com/willieneis/bayesian-algorithm-execution}}\footnote{This paper appears in \textit{Proceedings of the 38$^\text{th}$ International Conference on Machine Learning, 2021}.}
\end{abstract}
\vspace{-2mm}
\section{Introduction}
\label{sec:introduction}
\vspace{-1mm}

Many real-world problems can be described as inferring properties of an expensive
black-box function \(f\), subject to a computational budget of $T$ function evaluations.
This class of problems includes global optimization, commonly tackled by Bayesian
optimization methods \citep{shahriari2015taking, Frazier2018-bc}, but it also
encompasses many additional problems.
For example, in materials science, $f(x)$ may measure the strength of a material with
composition specified by \(x\in \Xcal\), and the goal might be to find the \emph{set}
of materials with strength above a threshold $C$, without ever evaluating more than
$T$ materials, due to the cost of such experiments
\citep{zhong2020accelerated, tran2021computational}.
Here, the property of interest is a set of points, the superlevel set of $f$.

Often, there exist effective algorithms to compute our property of interest
\emph{in the absence of a budget constraint.}
We call such a property a \emph{computable property} of $f$, if it is the output
of an algorithm $\Acal$ that makes a finite sequence of function evaluations
during its execution.
In the superlevel set example, $\Acal$ might simply evaluate $f$ at each $x\in \Xcal$
and output points with $f(x) > C$.
Other examples include using numerical quadrature to find integrals of $f$
\citep{davis2007methods}, using Newton's method to find roots of $f$ \citep{madsen1973root},
using evolution strategies or finite-difference gradient descent to find
local optima of $f$, and using Dijkstra's algorithm to find the shortest path between
nodes in a graph when the edge costs are given by $f$ \citep{dijkstra1959note}.
The property of interest in these examples take different forms, such as a single
value, a set of vectors, or a sequence of edges in a graph.
In each case, an algorithm for computing the property exists, but
executing that algorithm on $f$ may exceed our budget of $T$ evaluations.

In this paper, we address the general problem of estimating a computable property
$\algoutput := \algoutput(f)$ of a black-box function $f$, under a budget constraint $T$,
\emph{irrespective of the number of evaluations required by the algorithm $\Acal$}.
To do this, we posit a probabilistic model for $f$, and use it to estimate $\algoutput$
given data gathered via function evaluations.
Our goal is to make the best $T$ evaluations to yield an accurate estimate.
We refer to this problem as \textit{Bayesian algorithm execution}, or BAX.
Note that the probabilistic nature of BAX enables us to work with noisy function evaluations, e.g. $y_{x} \sim f(x) + \Ncal(0, \sigma^2)$, even if $\Acal$ is only designed for noiseless settings.

We develop an iterative procedure for BAX, called \infobaxnospace, that sequentially
evaluates the $x \in \Xcal$ that maximizes the mutual information (MI) between $f(x)$
and $\algoutput$ under our probabilistic model.
Each iteration of our procedure can be seen as an instance of Bayesian optimal experimental
design (BOED) where we choose an input to make an observation that maximally reduces the
uncertainty in the property of interest $\algoutput$ \citep{Chaloner1995-op}. However,
unlike a typical BOED setting, here the randomness in $\algoutput$ comes completely from
the uncertainty in $f$, and $\algoutput$ is generated by executing algorithm $\Acal$ on
$f$. Thus, we have neither access to a likelihood $p(y \given \algoutput, x)$ nor prior
$p(\algoutput)$, as is usually assumed in BOED, leading to computational challenges
that we address.

Our procedure relates to BOED methods for Bayesian optimization, such as entropy
search methods \citep{Hennig2012-zu, Hernandez-Lobato2014-ec, Wang2017-fb}, which leverage
a global optimization algorithm to compute a MI objective (our method can be viewed as an
extension of this branch of methods to computable function properties, beyond global
optima), and also to the MI criterion for optimal sensor placement via Gaussian
processes (GPs) \citep{Krause2008-zu}, which we can also view as estimating a
certain computable function property.
We discuss connections to these methods in Section~\ref{sec:background}.

All together, our method iteratively evaluates $f$ at $x \in \Xcal$ that maximally
reduces the uncertainty, measured by the posterior entropy, of the function property
at each step, and can therefore be used to estimate this property using minimal function
evaluations. In summary, our contributions are:
\begin{itemize}[parsep=0pt, topsep=5pt, itemsep=5pt]
    \item We introduce Bayesian algorithm execution (BAX), the task of inferring a computable
    property \(\algoutput\) of a black-box function \(f\) given an algorithm \(\Acal\) and a
    prior distribution on $f$, as a general framework that encapsulates many computational
    problems under uncertainty.
    \item We present an iterative, MI-maximizing procedure for BAX called \infobaxnospace, and
    present effective estimators of the MI objective that rely only on the ability to simulate
    $\Acal$ on posterior samples of $f$.
    \item We demonstrate the applicability of our methods in various settings, including for
    estimating graph properties (such as shortest paths) via Dijkstra's algorithm, and local
    optima (for variants of Bayesian optimization) via evolution strategies.
\end{itemize}
\section{Related Work}
\label{sec:background}

\paragraph{Bayesian optimal experimental design}
In BOED, we wish to estimate an unknown quantity or statistic \(\varphi\) through an
observation \(y_x\) resulting from an action or design \(x\). The goal is to choose
the design \(x\) that results in an observation \(y_x\) that is most informative
about the quantity of interest \(\varphi\). Typically, in BOED we assume that we have
access to an observation likelihood \(p(y_x \given \varphi)\) and a prior \(p(\varphi)\).
One popular strategy is then to maximize the expected information gain (EIG)
\citep{lindley1956measure} about \(\varphi\) from observing \(y_x\).
This is equivalent to the mutual information between \(\varphi\) and \(y_x\), which
can be written as
\begin{align}
    \label{eq:eig_boed_1}
    \EIG(x) = I(y_x, \varphi)
    = \mathbb{E}_{p(y_x | \varphi) p(\varphi)} \left[\log \frac{p(\varphi \given y_x)}{p(\varphi)} \right].
\end{align}
The Bayesian optimal design is then \(\argmax_x \EIG(x)\).
In practice, one often uses Monte Carlo or variational approximations of
this BOED objective \citep{Chaloner1995-op, Muller2005-lh, seeger2008large}.

Our setting is similar in structure to sequential BOED but differs in its assumptions
of what is computationally available to the practitioner.
For us, the unknown quantity $\varphi$ is the output of an algorithm $\algoutput$,
while $y_x$ are noisy observations of $f$ at inputs $x$.
We can neither compute the likelihood
$p(y_x \given \varphi) = p(y_x \given \algoutput)$ nor the prior
\(p(\varphi) = p( \algoutput) \), since we allow for arbitrary algorithms $\Acal$.
Furthermore, we cannot even sample from the likelihood for a given $\varphi = \algoutput$,
as in likelihood-free BOED \citep{drovandi2013bayesian, Kleinegesse2019-xg, Kleinegesse2020-rh}.
Recent work \citep[][see \textit{Extrapolation} example]{Foster2019-fj} has also considered
special cases of this setting.

\paragraph{BOED for function properties}
A number of prior works have presented BOED-based approaches for inferring specific function
properties using a probabilistic model for $f$, such as a Gaussian process.
Here we focus on two examples which relate closely to our framework: entropy search methods 
and optimal sensor placement.

Entropy search (ES) methods \citep{Hennig2012-zu, Hernandez-Lobato2014-ec, Wang2017-fb}
are Bayesian optimization procedures that can be viewed as BOED,
where the function property of interest is 
$\varphi := x^* := \argmax_{x \in \Xcal} f(x)$, the global optimizer of $f$.
Algorithms for ES typically operate on samples from the posterior distribution over $x^*$
(or its value, $f^* = f(x^*)$). To generate these samples, an optimization algorithm
is run on posterior samples of $f$, and the sampled outputs of this algorithm allow for
Monte Carlo estimates of the BOED MI objective $I(y_x, x^*)$ or $I(y_x, f^*)$, which is 
used as an acquisition function to choose subsequent $x_t$ to evaluate.
Below we will propose procedures for BAX that follow a similar strategy, and can be viewed as
extensions of ES methods to computable function properties beyond global optima.
We also note that an earlier black-box optimization method known as
Informational Approach to Global Optimization (IAGO) \citep{villemonteix2009informational} 
describes a similar objective as entropy search, albeit with a different computational
procedure to approximate this objective.

Another setting related to BOED is the sensor placement problem \citep{caselton1984optimal}.
Given budget of $T$ sensors and a set of potential sensor locations \(\Xcal\), we seek
to find \(X\subseteq \Xcal\) with $|X|=T$ that is ``most informative'' about the
measurement of interest, \(f: \Xcal \to \mathbb{R}\). When we measure
``informativeness'' by the mutual information between \(f(X)\) and \(f(\overline X)\),
the problem becomes NP-hard. In this setting, \citet{Krause2008-zu} proposed a
$1-1/e$ approximation algorithm that iteratively selects the sensor that maximizes
the \emph{gain} in mutual information. 
The sensor placement problem becomes a special case of BAX when we seek to infer the
value of $f$ at fixed locations \(X'\subseteq \Xcal\) and \(\Acal\) is the algorithm
that evaluates $f$ on each \(x \in X'\).

In addition, there has been work on using BOED methods with GP models to estimate a variety of
function properties, which are \textit{not} based on the above MI objective.
For example, the framework of stepwise uncertainty reduction on GPs has led to 
sampling objectives for tasks such as estimation of level sets or
excursion sets \citep{bect2012sequential, chevalier2014fast}. 
As another example, the framework of myopic posterior sampling (MPS) \citep{Kandasamy2019-qz}
takes a Thompson sampling-based approach, and has been applied to tasks such as active learning,
active posterior estimation, and level set estimation.
There have also been methods developed for estimating certain function properties in
tasks such as active surveying \citep{garnett2012bayesian}, quadrature \citep{osborne2012active},
sensor set selection \citep{garnett2010bayesian}, and
active search of patterns and regions \citep{ma2014active, ma2015active}.
\section{Bayesian Algorithm Execution (BAX)}
\label{sec:methodoverview}

In Bayesian algorithm execution (BAX), our goal is to estimate
$\algoutput := \algoutput(f) \in \Ocal$, the output of an algorithm $\Acal$ run on a
black-box function $f: \Xcal \rightarrow \Ycal$,
by evaluating $f$ on carefully chosen inputs $\{x_i\}_{i=1}^T\subseteq \Xcal$.
We will leverage a probabilistic model for \(f\) to guide our choice of $x$, 
in order to estimate $\algoutput$ using a minimal number of evaluations.

We assume that our initial uncertainty about the true function is captured by a
prior distribution over $f$, denoted by $p(f)$, reflecting our prior beliefs about $f$.
One notable example is the case where $p(f)$ is defined by a Gaussian process (GP).
Although not strictly necessary, we assume that each observation $y$ of the true function
$f_x := f(x)$ at input $x$ is noisy, with $y_x \sim f_x + \epsilon$ where
$\epsilon \sim \Ncal(0, \sigma^2)$.
We denote a dataset of $t-1$ function observations as
$\datasett = \{(x_i, y_{x_i})\}_{i=1}^{t-1}$, and use $p(f \given \datasett)$ to
denote the posterior distribution over $f$ given observations $\datasett$.
Given this distribution over $f$, and an algorithm $\Acal$ that returns as output
the computable property of interest $\algoutput$, we use $p(\algoutput \given \datasett)$
to denote the induced posterior distribution over the algorithm output.

\paragraph{Information-based BAX}
Under the above assumptions, we propose a sequential procedure to choose inputs
that are most informative about the property of interest $\algoutput$.
At each iteration $t$, we have a dataset of observations $\datasett$, and we
choose an input $x$ that maximizes the mutual information between $\algoutput$
and the unrevealed observation $y_x$.
The mutual information between two random variables $A$ and $B$ can be interpreted
as the expected information gain about $A$ upon observing $B$.
In our case, we choose $x$ to maximize this expected information gain about
$\algoutput$ given $y_x$, conditioned on our dataset $\datasett$, written
\begin{align}
    \label{eq:acqf_output_orig} 
    \EIG_t(x) =\hspace{1mm} \entropyf{\algoutput \given \datasett}
    - \expecf{p(y_x|\datasett)}
    { \entropyf{\algoutput \given \datasett \cup \{ (x, y_x) \} } }.
\end{align}
Here, $\entropyf{\algoutput \given \datasett} =
\expecf{p( f \given \datasett)}{-\log p( \algoutput \given \datasett)}$
is the entropy of $p(\algoutput \given \datasett)$,
and $p(y_x \given \datasett) = \expec{p(f \given \datasett)}{p(y_x \given f)}$
denotes the posterior predictive distribution at input $x$ after observing data
$\datasett$. In the following subsections, we will focus on developing practical
methods to estimate $\acqfnt(x)$.

Similar to methods in Bayesian optimization and sequential BOED, our full procedure is as follows.
At each iteration $t$, we use $\acqfnt(x)$ as an  acquisition function.
We optimize this acquisition function to choose the next input to query, i.e.
$x_t \leftarrow \argmax_{x \in \Xcal} \acqfnt(x)$.
We then evaluate the function $f$ at $x_t$ to observe a value
$y_{x_t} \sim f_{x_t} + \epsilon$, and update our dataset
$\dataset_{t+1} \leftarrow \datasett \cup \{(x_t, y_{x_t})\}$, before continuing to iteration $t+1$.
We refer to this procedure as \infobax (Algorithm~\ref{alg:infobax}).

\paragraph{Algorithm execution path}
Suppose that when we execute algorithm $\Acal$ on $f$ to compute $\algoutput$,
there are $S$ function evaluations.
We refer to the \emph{sequence} of function inputs and outputs traversed during
the execution of the algorithm as the \emph{execution path of $\Acal$ on $f$},
denoted  $\exepath := \exepath(f) := (z_s, f_{z_s})_{s=1}^S$.

Note that each input $z_s$ in the execution path may depend on all previous
inputs and outputs, e.g. 
$z_2 := z_2(z_1, f_{z_1})$,
$z_3 := z_3(z_1, f_{z_1}, z_2, f_{z_2})$, and 
$z_s$ $:=$ $z_1(z_1, f_{z_1}, \ldots, z_{s-1}, f_{z_{s-1}})$ in general.
For example, algorithm $\Acal$ may have specifically queried $z_2$ during its
execution because it observed the value $f(z_1)$ at $z_1$. 
To highlight the fact that inputs on the execution path have these dependencies,
we use the notation $z$, instead of $x$.
Likewise, we note that the output of $\Acal$ is a function of the
execution path, i.e.
$\algoutput(f)
:= \algoutput(\exepath(f))
= \algoutput \left( (z_s, f_{z_s})_{s=1}^S \right)$.

We will make use of this notion of \emph{execution paths} to define procedures for
computing $\EIG_t(x)$.
However, we emphasize that our procedures will not run $\Acal$ on the true $f$.
Instead, we will only run \(\Acal\) on function \emph{samples} $\samp{f}$ from
$p(f \given \datasett)$.

\begin{algorithm}[t]
    \caption{\textsc{InfoBAX}}
    \label{alg:infobax}
    \hspace*{\algorithmicindent}\hspace{1pt}\textbf{Input:} dataset $\dataset_1$, distribution \(p(f)\), algorithm \(\Acal\)
    \begin{algorithmic}[1]
      \For{$t = 1,\ldots,T$}
        \State $x_t \leftarrow \argmax_{x \in \Xcal} \acqfnt(x)$
            \Comment{See Equation (\ref{eq:acqf_output_orig})}
        \State $y_{x_t} \sim f(x_t) + \epsilon$
            \Comment{Evaluate $f$ at $x_t$}
        \State $\dataset_{t+1} \leftarrow \datasett \cup \{(x_t, y_{x_t})\}$
            \Comment{Update dataset}
      \EndFor 
    \end{algorithmic}
    \hspace*{\algorithmicindent}\hspace{1pt}\textbf{Output:} distribution $p(\algoutput(f) \given \dataset_{T+1})$
\end{algorithm}

\paragraph{Example: top-$k$ estimation}
Here we introduce a running example that will be used to help illustrate our methods.
Suppose we have a finite collection of elements $X \subseteq \Xcal$, where each
$x \in X$ has an unknown value $f_x$.
There are various applications where we care about estimating the
\textit{top-$k$ elements of $X$} with the highest values, denoted $K^* \subseteq X$.
For instance, each \(x \in X\) could be a candidate formula for concrete
with tensile strength \(f_x\), and we wish to find the top \(k\)
formulae with the highest strengths.
Note that if our budget of evaluations satisfies $T \geq |X|$, we could run the following
\textit{top-$k$ algorithm} $\Acal$:
evaluate  $f$ on each $x \in X$, sort $X$ in decreasing order, and return the 
first $k$ elements.
In contrast, since $T < |X|$, our goal will be to choose the best
$T$ inputs $x_1, \ldots, x_T$ to query, in order to infer $K^*$.
For full generality, assume that we can evaluate any $x_t \in \Xcal$,
so we are not restricted to evaluating only inputs in $X$.

\begin{figure}[t]
\centering
\includegraphics[width=.68\textwidth]{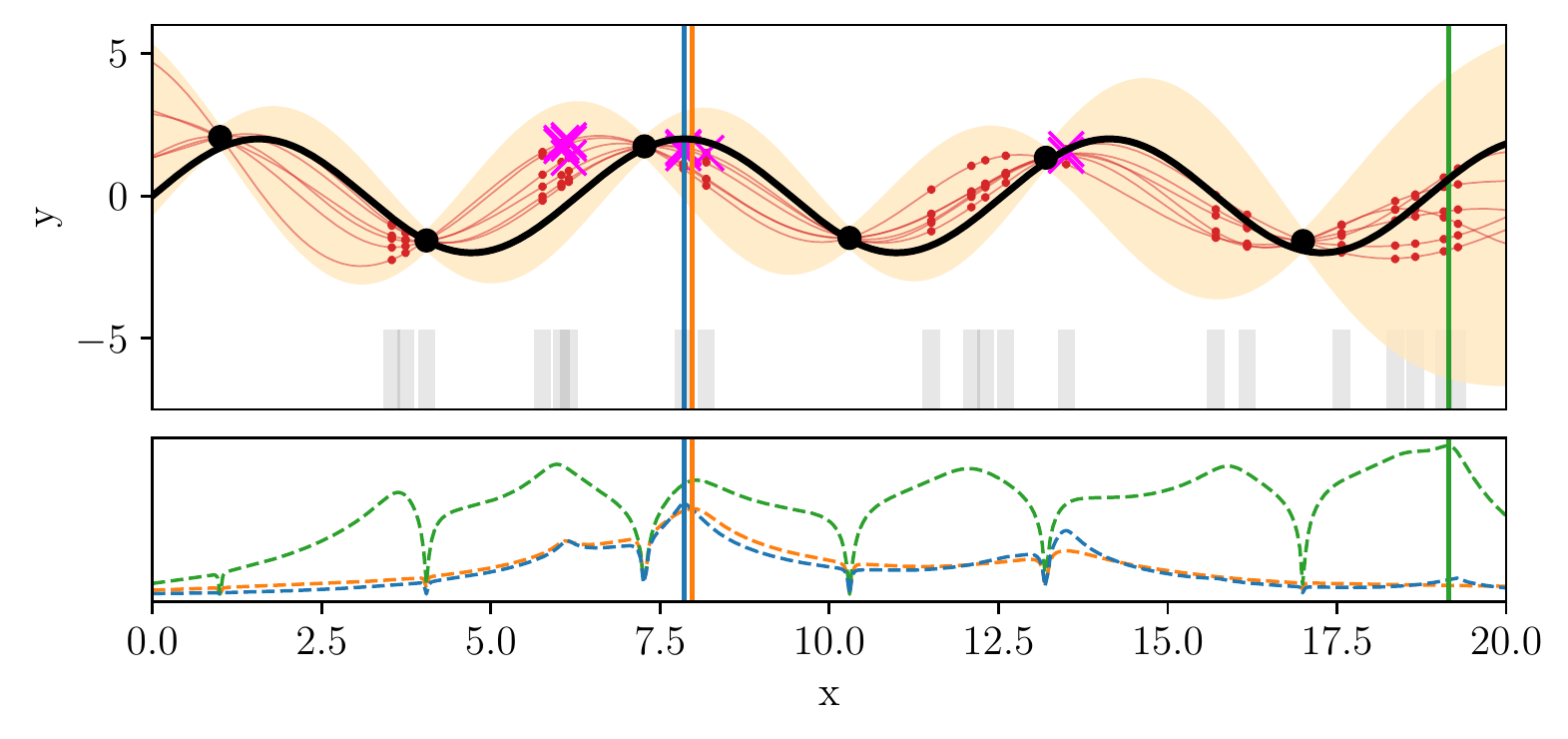}\\
\;\;\;\includegraphics[width=0.6\textwidth]{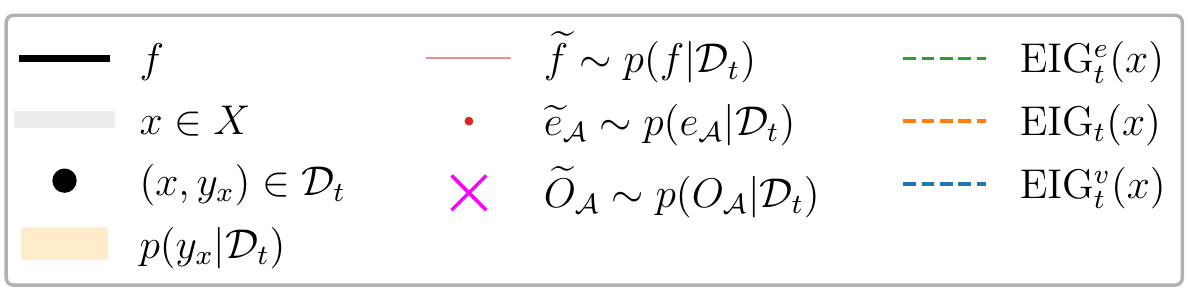}
\caption{
Illustrations of \textsc{InfoBAX} acquisition functions.
Here, $\Acal$ is the \textit{top-$k$ algorithm} on a set of elements $X$,
for $k=2$ (see text for description).
We show the function $f$, elements $x \in X$, observations $(x, y_x) \in \datasett$,
posterior predictive distribution $p(y_x \given \Dcal_t)$,
posterior samples (of the function $\samp{f}$, execution path $\exepathsamp$, and 
algorithm outputs $\algoutputsamp$), and $\EIG$ acquisition functions 
(\ref{eq:acqf_execpath}), (\ref{eq:acqf_output_expect}), and (\ref{eq:acqf_subexecpath}).
The three vertical lines denote the $\argmax$ of the three acquisition functions.
}
\label{fig:topk}
\end{figure}

Under algorithm \(\Acal\), the execution path $\exepath$ has a fixed sequence of inputs
$(z_1, \ldots, z_{|X|})$ equal to an arbitrary ordering of the $x \in X$.
Given a distribution  \(p(f \given \Dcal_t)\) over the function $f$ 
conditioned on some observations \(\Dcal_t\), we can estimate the top-\(k\) elements
by executing \(\Acal\) on samples \(\samp{f} \sim p(f \given \Dcal_t)\).
We illustrate this procedure in Figure~\ref{fig:topk} for \(k=2\).
Here, the set $X$ is shown as a set of short grey bars.
We also show the true function $f$ (black line), six observations
$(x, y_x) \in \datasett$ (black dots), the posterior predictive distribution
$p(y_x \given \Dcal_t)$ (tan shaded band), and samples $\samp{f}$ (red lines).

\paragraph{Summary of acquisition functions}
In the following sections, we present three acquisition functions for
\infobaxnospace, which approximate Eq.~(\ref{eq:acqf_output_orig}).
First, in Section~\ref{sec:eigexepath}, we introduce an acquisition
function which is in general suboptimal; however, it will help us define and
describe how to compute the latter acquisition functions.
After, in Section~\ref{sec:eigoutput}, we present an acquisition 
function which is the optimal quantity that we want,
but may be computationally costly to compute.
Finally, in Section~\ref{sec:eigexepathsubsequence}, we present an acquisition
function that approximates the previous one and is much cheaper to compute,
but comes with some restrictions on settings where it should be used.

\subsection{EIG for the Execution Path}
\label{sec:eigexepath}

As a first step toward developing our method, we will show how to compute a modified
$\EIG$ objective.
Note that the execution path $\exepath$ is a sequence of $(z_s, f_{z_s})$ pairs which,
taken together, give complete information about the algorithm's output---meaning that if we
knew the function value at each point in $\exepath$, then we would know the function property
$\algoutput$ exactly.
Consequently, one potential strategy for BAX is to, at each iteration, choose to query the
$x \in \Xcal$ that gives most information about the execution path (i.e. maximally reduce
the entropy of the distribution over $\exepath$).

We therefore first present a modified version of the acquisition function $\acqfnt(x)$
in (\ref{eq:acqf_output_orig}).
Let $\acqfnt^e(x)$ be the expected gain in information on the execution path $e_\Acal$,
written
\begin{align}
    \label{eq:acqf_execpath_orig}
    \acqfnt^e(x) = \hspace{1mm} \entropyf{\exepath \given \datasett}
    - \expecf{p(y_x | \datasett)}
    { \entropyf{\exepath \given D_t \cup \{(x, y_x)\} } }.
\end{align}
In the case where the property $\algoutput = \exepath$, this acquisition function is
equal to (\ref{eq:acqf_output_orig}), i.e. $\acqfnt^e(x) = \acqfnt(x)$.
Otherwise, the two acquisition functions are distinct in general.

There are a few difficulties in computing $\acqfnt^e(x)$ as it is written in
(\ref{eq:acqf_execpath_orig}).
For the first term, we must estimate the entropy of $p(\exepath \given \datasett)$, which
is analytically intractable and potentially very high dimensional.
If we did have a way to estimate this entropy, we could do the following for the second
term for a given $x\in \Xcal$: sample a set of $\samp{y}_x \sim p(y_x \given \datasett)$,
and for each $\samp{y}_x$ sample, re-train our model on $\datasett \cup \{(x, \samp{y}_x)\}$
and estimate the entropy $\entropyf{\exepath \given \datasett \cup\{(x,\samp{y}_x)\}}$ using
the same technique used for the first term. These steps are expensive, and would need to be
repeated for each $x \in \Xcal$ over which we intend to optimize our acquisition function.

Instead, we follow an approach from prior work \citep{Hernandez-Lobato2014-ec, Houlsby2012-xf}.
Since (\ref{eq:acqf_execpath_orig}) is the MI between $\exepath$ and $y$ (given $\datasett$),
and due to the symmetry of MI, we can rewrite this acquisition function as
\begin{align}
    \label{eq:acqf_execpath}
    \acqfnt^e(x) = \hspace{1mm} \entropyf{y_x \given \datasett}
    - \expecf{p( \exepath | \datasett)}
    { \entropyf{ y_x \given \datasett, \exepath} }.
\end{align}
The first term is simply the entropy of the posterior predictive distribution
$p(y_x \given \datasett)$, which we can compute exactly for certain probabilistic
models such as GPs. For the second term, inside the expectation, we have
$\entropyf{y_x \given \datasett, \exepath}$, which is the entropy of the posterior
predictive distribution at input $x$, given both the dataset $\datasett$ and the
execution path $\exepath$.

Before explaining how to compute $p\left( y_x \given \datasett, \exepath \right)$
and its entropy for a given $\exepath$, we first describe how to compute the
expectation of this entropy with respect to $p(\exepath \given \datasett)$.
To do this, we will compute a Monte Carlo estimate via a Thompson sampling-like strategy,
related to procedures used by entropy search methods
\citep{Hennig2012-zu, Hernandez-Lobato2014-ec, Wang2017-fb}.
We first draw $\samp{f} \sim p(f \given \datasett)$, and then run our algorithm
$\Acal$ on $\samp{f}$ to produce a sample execution path $\exepathsamp = \exepath(\samp{f})$.
Note that this yields a sample $\exepathsamp \sim p(\exepath \given \datasett)$.
In Section~\ref{sec:computational}, we give details on how we implement this procedure for
GP models in a computationally efficient manner.

We repeat this multiple times to generate a set of $\ell$ posterior execution path samples
$\{\exepathsamp^{\;j}\}_{j=1}^\ell$.
Notably, unlike the procedure for Eq.~(\ref{eq:acqf_execpath_orig}), we only need to perform
this sampling procedure once, and then can use the same set of samples to compute
$\acqfnt^e(x)$ for all $x \in \Xcal$. Concretely, to compute $\acqfnt^e(x)$,
we compute $\entropy{y_x \given \datasett, \exepathsamp^{\;j\;}}$ for each sample
$\exepathsamp^{\;j\;}$, and average these to form a Monte Carlo estimate of the second term in
Eq.~(\ref{eq:acqf_execpath}), i.e. with
$\frac{1}{\ell} \sum_{j=1}^\ell \entropy{y_x \given \datasett, \exepathsamp^{\;j\;}}$.

We now describe how to compute $\entropy{y_x \given \datasett, \exepathsamp}$.
The key idea is that, under our modeling assumptions, we can derive a closed-form expression for
$p(y_x \given \datasett, \exepathsamp)$ in which we can compute the entropy analytically.
Let the posterior execution path sample $\exepathsamp$ be comprised of the sequence
$\exepathsamp = \big( \samp{z}_s, \samp{f}_{z_s} \big)_{s=1}^{S}$.
We can then show that
\begin{align}
    \label{eq:postpred_given_exepath}
    p \left(y_x \given \datasett, \exepathsamp \right)
    = p \left( y_x \; \Big| \;
    \datasett, \big\{ \samp{f}_{z_s} \big\}_{s=1}^S \right).
\end{align}
This is equivalent to computing the posterior predictive distribution,
given observations with different noise levels, where observations
$\datasett$ are assumed to have noise given by the likelihood, and variables
$\samp{f}_{z_s}$, are treated as noiseless observations.
Under our GP model, this can be computed exactly in closed form (it is a Gaussian distribution),
as can $\entropy{y_x \given \datasett, \exepathsamp}$.
We show this and give the explicit formula for the GP model in Section~\ref{sec:eigexepathappendix}.

In Figure~\ref{fig:topk}, we show this acquisition function as the green dashed line.
We also show the execution path samples $\exepathsamp^{\;j\;} = \exepath(\samp{f}^{\;j\;})$,
used to compute $\EIG_t^e$, as red dots over posterior function samples $\samp{f}^{\;j}$ (red lines).

We note again that when $\algoutput \neq \exepath$, using $\acqfnt^e(x)$ in
Eq.~(\ref{eq:acqf_execpath}) may be effective in practice, but is suboptimal.
For example, given an algorithm where a subsequence of the execution path has no
influence on later parts of the execution path nor on the algorithm output, by
following the above procedure we may waste queries on estimating portions of
$\exepath$ that do not give much information about $\algoutput$.

\subsection{EIG for the Algorithm Output}
\label{sec:eigoutput}

We next show how to use the equations derived above to compute the
expected information gain on the algorithm output $\algoutput$.
First, we rewrite the acquisition function $\acqfnt(x)$ from Eq.~(\ref{eq:acqf_output_orig}) in a
predictive entropy-based form (analogous to what was done in Eq.~(\ref{eq:acqf_execpath})), i.e.
\begin{align}
    \label{eq:acqf_output}
    \acqfnt(x) = \hspace{1mm} \entropyf{ y_x \given \datasett } 
    - \expecf{p( \algoutput | \datasett)}
    { \entropyf{ y_x \given \datasett, \algoutput } }.
\end{align}
Unlike the previous strategy, it is difficult to compute
$p(y_x \given \datasett, \algoutput)$ in Eq.~(\ref{eq:acqf_output}),
in general, for any algorithm $\Acal$,
due to conditioning on the algorithm output $\algoutput$.
This distribution is the posterior predictive at an input $x$
given dataset $\datasett$, and \textit{also conditioned on} the black-box function having property $\algoutput$.
While we can compute $p(y_x \given \datasett, \exepath)$ in closed form under certain models, 
since $\exepath$ is a sequence of inputs and function values,
this is not the case when we condition on $\algoutput$, which can be an arbitrary property
of $f$. However, by using the execution path as an auxiliary variable, we can equivalently
write this posterior as
\begin{align}
    p(y_x \given \datasett, \algoutput)
    &= \int p(y_x, \exepath \given \datasett, \algoutput) \diff\exepath \nonumber \\
    &= \int p(y_x \given \datasett, \exepath, \algoutput) \hspace{1mm}
    p(\exepath \given \algoutput, \datasett) \diff\exepath  \nonumber \\
    &= \expecf{p(\exepath | \algoutput, \datasett)}{
        p \left( y_x \given \datasett, \exepath \right)
    }.
\end{align}
Here we use the fact that $\Acal$ defines an execution path that specifies
the algorithm output exactly, and thus $y \bigCI \algoutput \given \exepath$.
We can therefore write Eq.~(\ref{eq:acqf_output}) as
\begin{align}
    \label{eq:acqf_output_expect}
    \acqfnt(x) = \hspace{1mm}
    \entropyf{ y_x \given \datasett}
    - \expecf{p( \algoutput | \datasett)}
    {
        \entropyf{\expecf{p(\exepath | \algoutput, \datasett)}{
        p \left( y_x \given \datasett, \exepath \right)}}
    }.
\end{align}

Given that we have access to $p \left( y_x \given \datasett, \exepath \right)$ in 
closed form, we can estimate the expression
$\expecf{p(\exepath \given \algoutput, \datasett)}{p \left( y_x \given \datasett, \exepath \right)}$
using
$\frac{1}{\ell}\sum_{k=1}^\ell p \left( y_x \given \datasett, \exepathsamp^{\;k\;} \right)$,
where $\exepathsamp^{\;k\;}$ $\stackrel{iid}{\sim}$ $p(\exepath \given \algoutput, \datasett)$.
By sampling from this, we can approximate the entropy in Eq.~(\ref{eq:acqf_output_expect})
via a Monte Carlo estimate. Therefore, the key question is how to draw samples from
$p(\exepath \given \algoutput, \datasett)$.
Intuitively, a sample $\exepathsamp^{\;k\;} \sim p(\exepath \given \algoutput, \datasett)$
is a plausible execution path, given observations $\datasett$, which also yields output $\algoutput$.
At a given iteration of \textsc{InfoBAX}, suppose we generate a set of samples from the posterior
over algorithm outputs, $\{ \algoutputsamp^{\;j\;} \}_{j=1}^\ell \stackrel{iid}{\sim}
p(\algoutput \given \datasett)$ by running $\Acal$ on posterior function samples $\samp{f}^{\;j\;}$.
Suppose also that we have defined a distance $d(\cdot, \cdot)$
on our algorithm output space $\Ocal$. For each $\algoutputsamp^{\;j\;}$,
we could then define a \textit{set of similar outputs} $\algoutputring^{\;j\;}$ to be
\begin{align}
\algoutputring^{\;j\;} = 
\left\{
\outputplainsamp \in \{ \algoutputsamp^{\;k\;} \}_{k=1}^\ell
\;:\;
d (\outputplainsamp, \algoutputsamp^{\;k\;} ) < \delta, 
\hspace{1mm} k \neq j
\right\}, \nonumber
\end{align}
i.e. all outputs within a ball of diameter $\delta$ centered at $\algoutputsamp^{\;j\;}$.
Intuitively, we can then compute the EIG on a ball of diameter $\delta$
in the output space that contains the algorithm output, rather than
on the algorithm output directly.

More formally, this can be viewed as an instance of approximate Bayesian computation (ABC)
\citep{beaumont2002approximate, csillery2010approximate}, 
which is a technique for generating posterior samples, given only a simulator for the likelihood.
In our case, by running $\Acal$, we can simulate an output $\algoutput$ given an execution path
$\exepath$, and use this to produce approximate posterior samples
from $p(\exepath \given \algoutput, \datasett)$.
Concretely, suppose we've sampled a set of pairs
$P^j := \{(\exepathsamp^{\;j\;}, \algoutputsamp^{\;j\;})\}$ by running algorithm $\Acal$
on samples $\samp{f} \sim p(f \given \datasett)$.
For each $\algoutputsamp^{\;j\;}$, we can then treat 
$\exepathring^{\;j\;} = \{ e \in P^j : \algoutputsamp \in \algoutputring^{\;j\;} \}$ 
as approximate samples from $p(\exepath \given \algoutputsamp^{\;j\;}, \datasett)$.
This is equivalent to the ABC algorithm from \citet{rubin1984bayesianly} and
\citet{beaumont2010approximate}. We then use the set of sample execution paths
$\exepathring^{\;j\;}$ to construct the sample estimate of $\acqfnt(x)$ in
(\ref{eq:acqf_output_expect}). We give explicit formulae for
(\ref{eq:acqf_output_expect}) under GP models in Section~\ref{sec:eigoutputappendix}.

As \textsc{InfoBAX} progresses, and we have better estimates of the algorithm output,
we can reduce the diameter $\delta$ and continue to yield large enough sample sets
$\exepathring^{\;j\;}$ to form accurate Monte Carlo estimates of
(\ref{eq:acqf_output_expect}). In practice, we choose $\delta$ to be
the smallest value such that every $\exepathring^{\;j\;}$ has size greater
than a fixed number (such as 30).

In Figure~\ref{fig:topk}, we show this acquisition function as the yellow dashed line.
We also show samples of the algorithm output $\algoutputsamp^{\;j\;}$
(from which we then produce $\algoutputring^{\;j\;}$) as magenta crosses.

\subsection{EIG using an Execution Path Subsequence}
\label{sec:eigexepathsubsequence}

One disadvantage of using $\acqfnt(x)$ in Eq.~(\ref{eq:acqf_output_expect}) is that it may
require a large set of samples
$\{\exepathsamp^{\;j}\}_{j=1}^\ell$ $\stackrel{iid}{\sim}$
$p(\exepath \given \algoutput, \datasett)$, 
in order to compute an accurate Monte Carlo estimate.
Instead, one final strategy we can attempt is to determine a latent variable $\subexepathplain$,
in which
\begin{enumerate}[nolistsep, topsep=2mm, label=(\roman*), leftmargin=12mm]
    \item we can draw samples $\subexepathplainsamp \sim p(\subexepathplain \given \datasett)$,
    \item we can compute $p \left( y_x \given \datasett, \subexepathplainsamp \right)$, and 
    \item the EIG with respect to $\subexepathplain$, $\acqfnt^v(x) \approx \acqfnt(x)$.
\end{enumerate}

\noindent A potential idea is to try and define a mapping from an execution path
$\exepath$ to a $\subexepathplain$ that best fits the above criteria.
For example, consider a subsequence of $\exepath$ of length $R$,
denoted $\subseqexepath := \subseqexepath(f) := (z_{i_r}, f_{z_{i_r}})_{r=1}^R$.
We can denote the \textit{function values} for this subsequence with
$\subexepath := \subexepath(f)$ $:=$ $\{f_{z_r}\}_{r=1}^R$, and then write
\begin{align}
    \label{eq:acqf_subexecpath}
    \acqfnt^v(x) = 
    \hspace{1mm} \entropyf{y_x \given \datasett }
    - \expecf{p( f | \datasett)}
    { \entropyf{ y_x \given \datasett, \{f_{z_r}\}_{r=1}^R } }.
\end{align}

Note that the posterior
$p \left( y_x \given \datasett, \subseqexepath \right)$ $\neq$
$p \left( y_x \given \datasett, \subexepath \right)$. 
The former, in general, depends on unconditioned latent variables in
the execution path $\exepath$, and is intractable to compute
(this is \textit{not} the case, however, when $\subseqexepath = \exepath$, as
 we show in (\ref{eq:postpred_given_exepath})).
On the other hand, for models such as GPs,
$p \left( y_x \given \datasett, \subexepath \right)$ can indeed
be computed exactly and its entropy available in closed form.
Hence, $\subexepath$ satisfies (ii).

Furthermore, to compute samples $\subexepathsamp \sim p(\subexepath \given \datasett)$
we can easily sample $\samp{f} \sim p(f \given \datasett)$, and then set
$\subexepathsamp = \subexepath(\samp{f})$, so $\subexepath$ satisfies (i) as well.
Note that, since we can sample $\subexepathsamp$ and compute
$p\left( y_x \given \datasett, \subexepathsamp \right)$, we can
estimate $\acqfnt^v(x)$ via a Monte Carlo estimate similar to~(\ref{eq:acqf_execpath}).

However, we still need to show that $\subexepath$ satisfies (iii).
For this, we focus on a special case of interest.
In some problems, the 
function property $\algoutput$ exactly specifies some function
values $\subexepath$ along a subsequence $\subseqexepath$ of the execution path. 
A few examples of such properties include optima (where $\subseqexepath$ consists of an
optima $x^*$ and its value $f_{x^*}$),
level sets (where $\subseqexepath$ is the set of $(x, f_x)$ pairs in a super/sublevel set),
function roots (where $\subseqexepath$ is a root of $f$), and phase boundaries 
(where $\subseqexepath$ is a set of $(x, f_x)$ pairs that comprise the phase boundary).
In these cases, for a given sample $\samp{f} \sim p(f | \datasett)$ with
associated $\algoutputsamp$ and $\subexepathsamp$,
we have that $p(y_x \given \datasett, \algoutputsamp)$ $=$
$p(y_x \given \datasett, \subexepathsamp, \algoutputsamp)$, and
\begin{align}
    p( y_x \given \datasett, \subexepathsamp) =
        \expecf{p(\algoutput | \subexepathsamp, \datasett)}{
            p(y_x \given \datasett, \subexepathsamp, \algoutput)
        } \nonumber
\end{align}
(see Section~\ref{sec:eigsubexeappendix} for details).
$\acqfnt^v(x)$ will thus serve as a better approximation when 
$\entropyf{\algoutput | \subexepath, \datasett}$ is small, and will
be optimal when $\entropyf{\algoutput | \subexepath, \datasett} = 0$,
in which case $\acqfnt^v(x) = \acqfnt(x)$.

Empirically, we often observe this behavior.
For example, in Figure~\ref{fig:topk}, we show $\acqfnt^v(x)$ as the blue dashed line,
which closely approximates $\acqfnt(x)$ (the yellow dashed line).
In cases such as those given above, where property $\algoutput$
specifies some function values $\subexepath$ along a subsequence of $\exepath$,
a computationally attractive and practically effective strategy is to use the
acquisition function $\acqfnt^v(x)$ in (\ref{eq:acqf_subexecpath}).
\section{Experiments}
\label{sec:experiments}
We evaluate our proposed \infobax method for Bayesian algorithm execution on three
applications in distinct domains. Our experiments demonstrate the generality of the
BAX framework for formalizing the task of inferring black-box function properties and
the effectiveness of \infobax for estimating graph properties, local optima, and top-$k$ sets.
In each problem, we use a property-computing algorithm $\Acal$ that was \emph{not} designed
for settings where we have a limited budget of function evaluations. Nevertheless, our
\infobax procedure lets us apply such algorithms under a budget constraint, allowing us to
infer the true algorithm output using significantly fewer queries than the algorithm alone
would have required.

We use Gaussian processes as our prior distribution $p(f)$ for all tasks. To reduce
computation time of posterior sampling, we use the sampling method proposed by
\citep{wilson2020efficiently} implemented in GPFlow \citep{matthews2017gpflow} with
GPU acceleration. We refer the reader to Section \ref{sec:experimentsappendix} for
additional details on our experimental setup as well as empirical comparisons
of our proposed MI objectives.

\begin{figure*}
\centering
\includegraphics[height=1.9in]{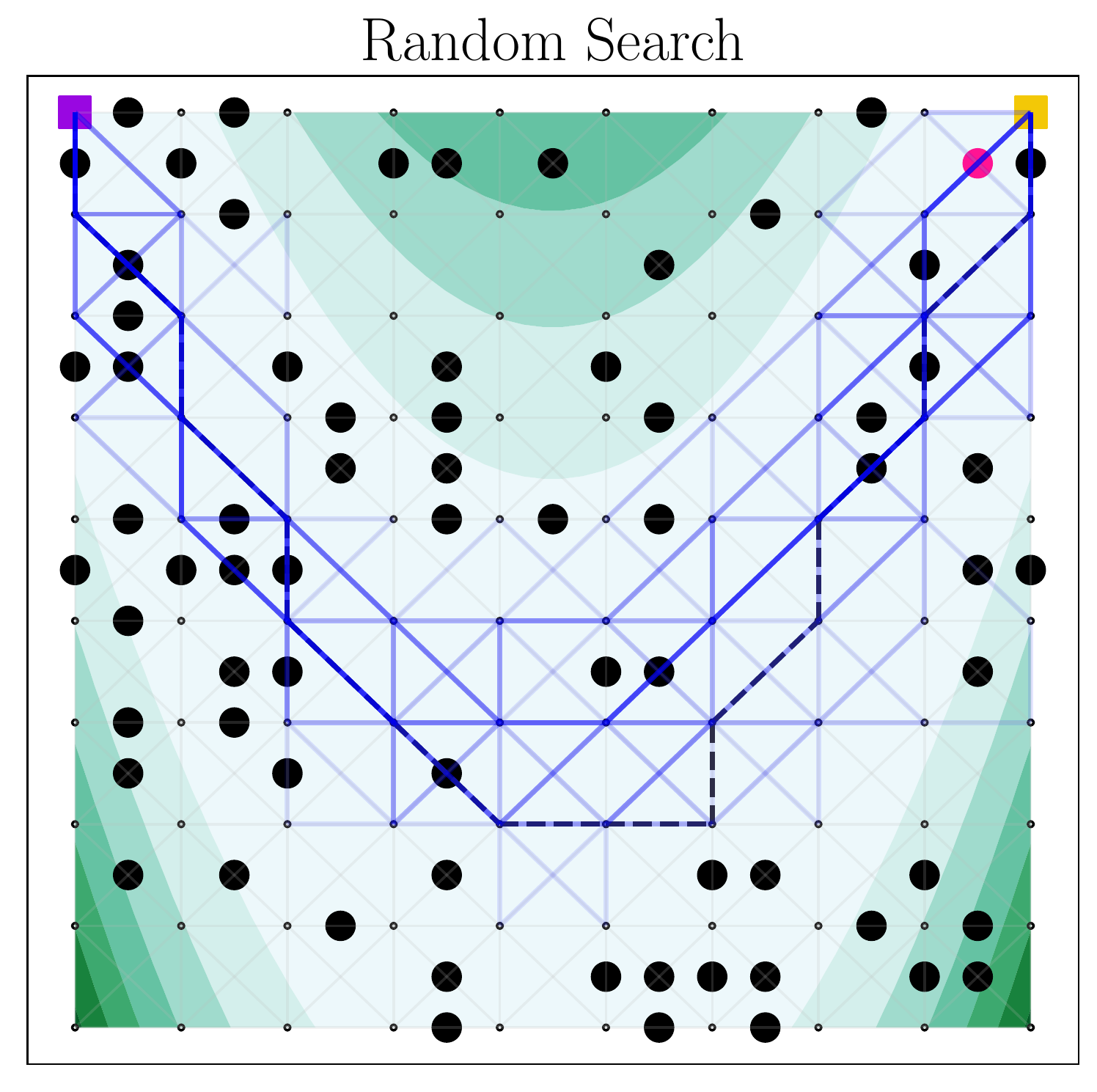}
\hspace{1mm}
\includegraphics[height=1.9in]{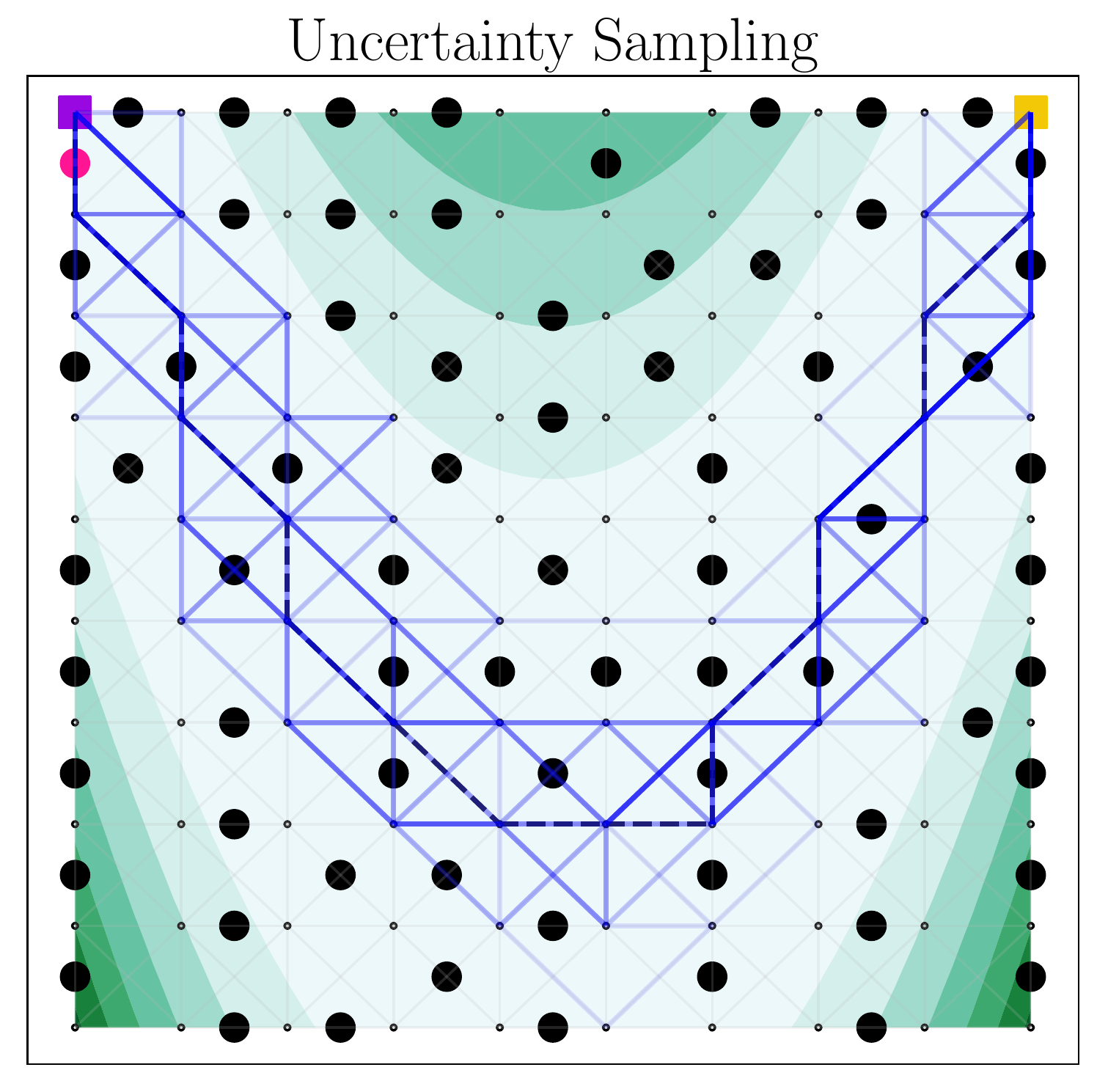}
\hspace{1mm}
\includegraphics[height=1.9in]{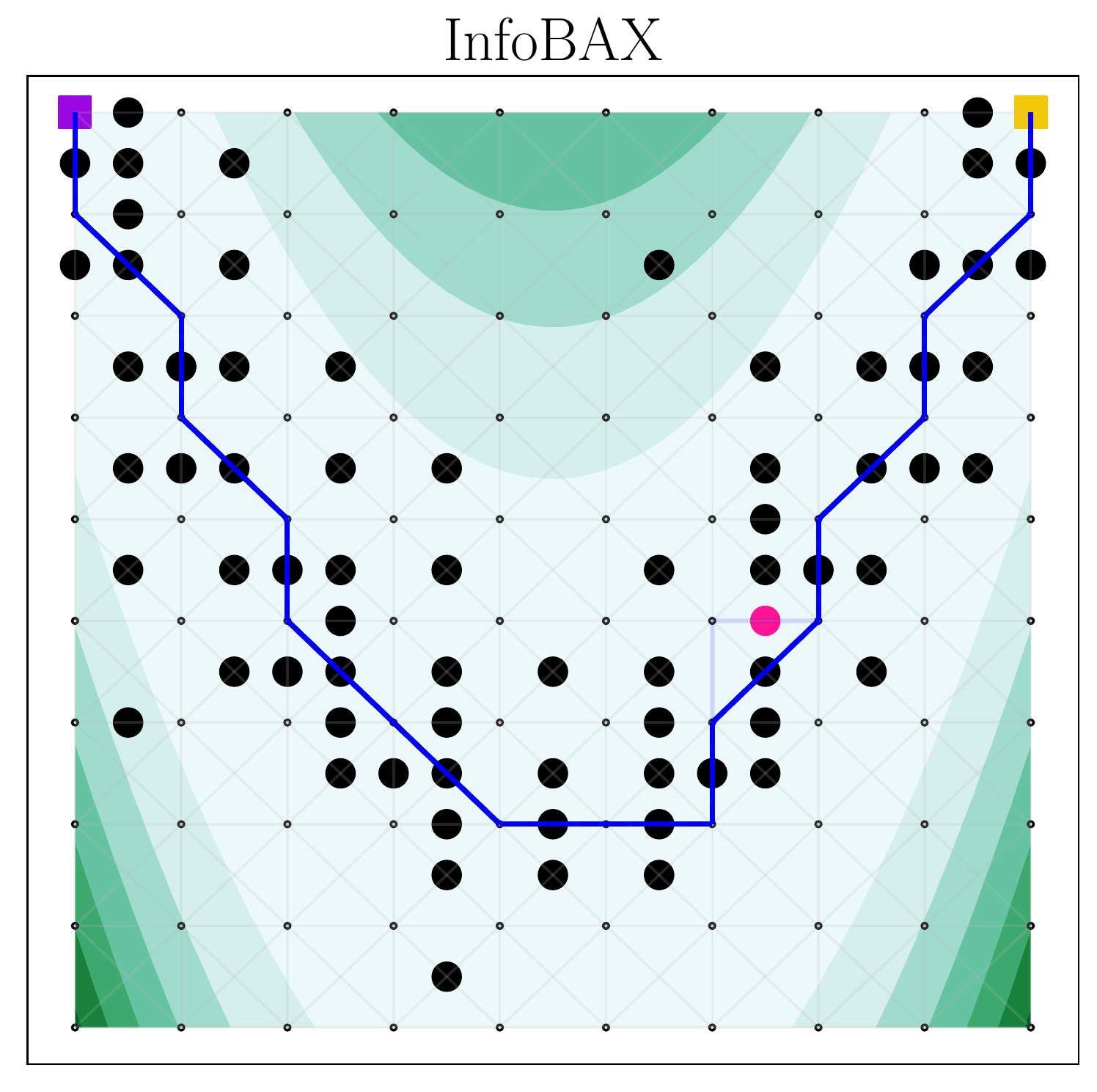}\\
\hspace{6.0mm} \includegraphics[height=2.0in]{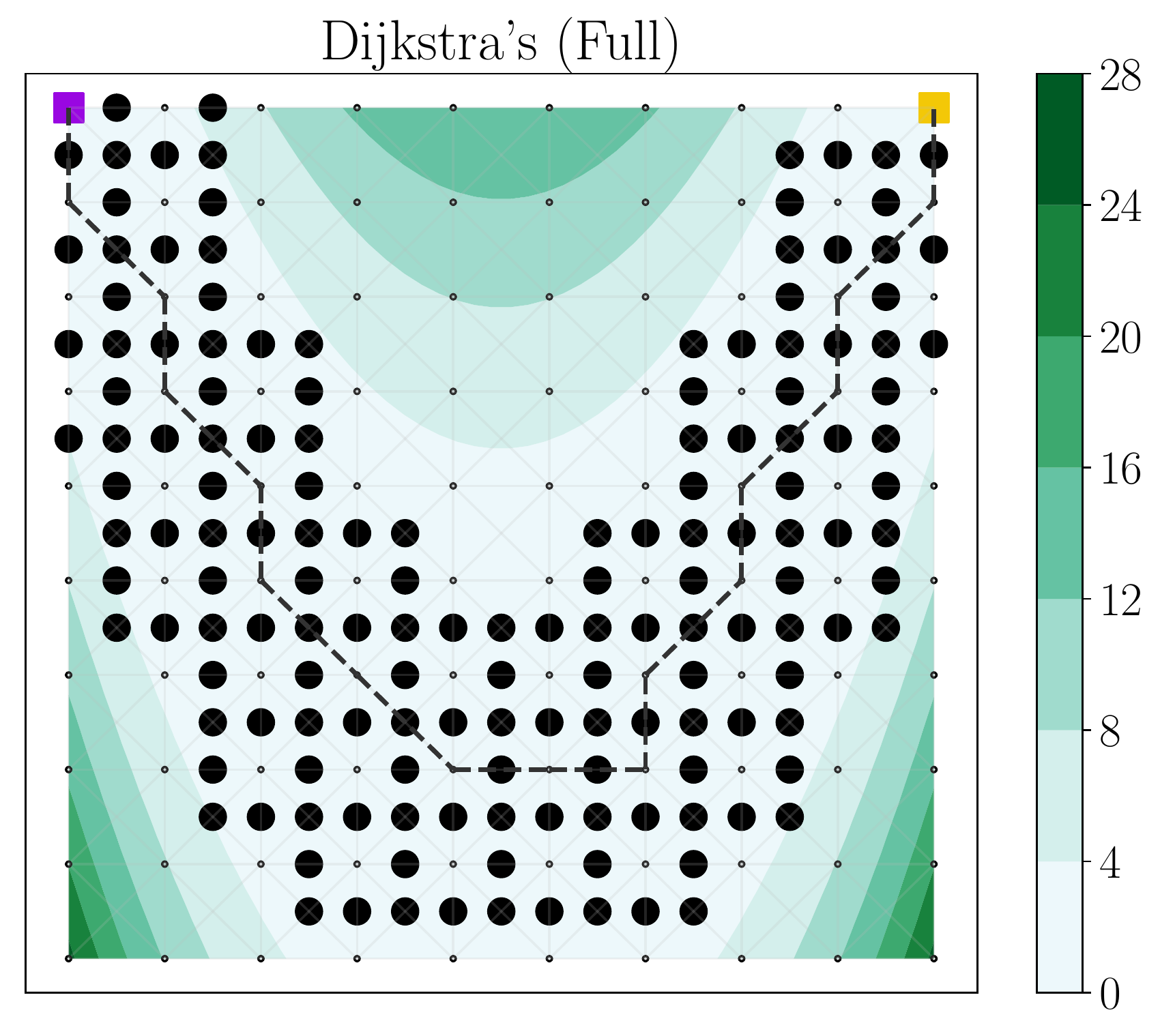}\\
\includegraphics[height=1.9in]{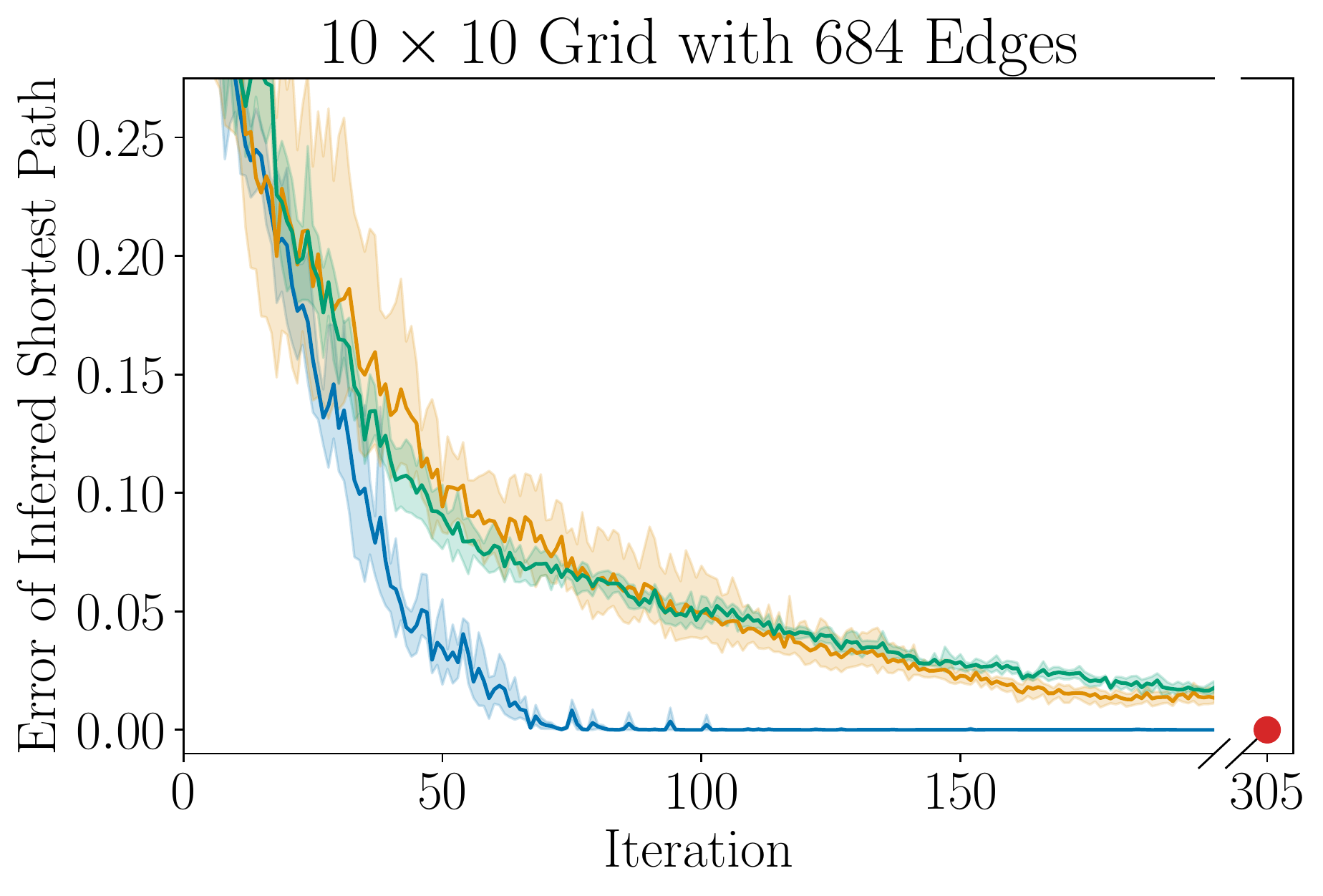}
\includegraphics[height=1.9in]{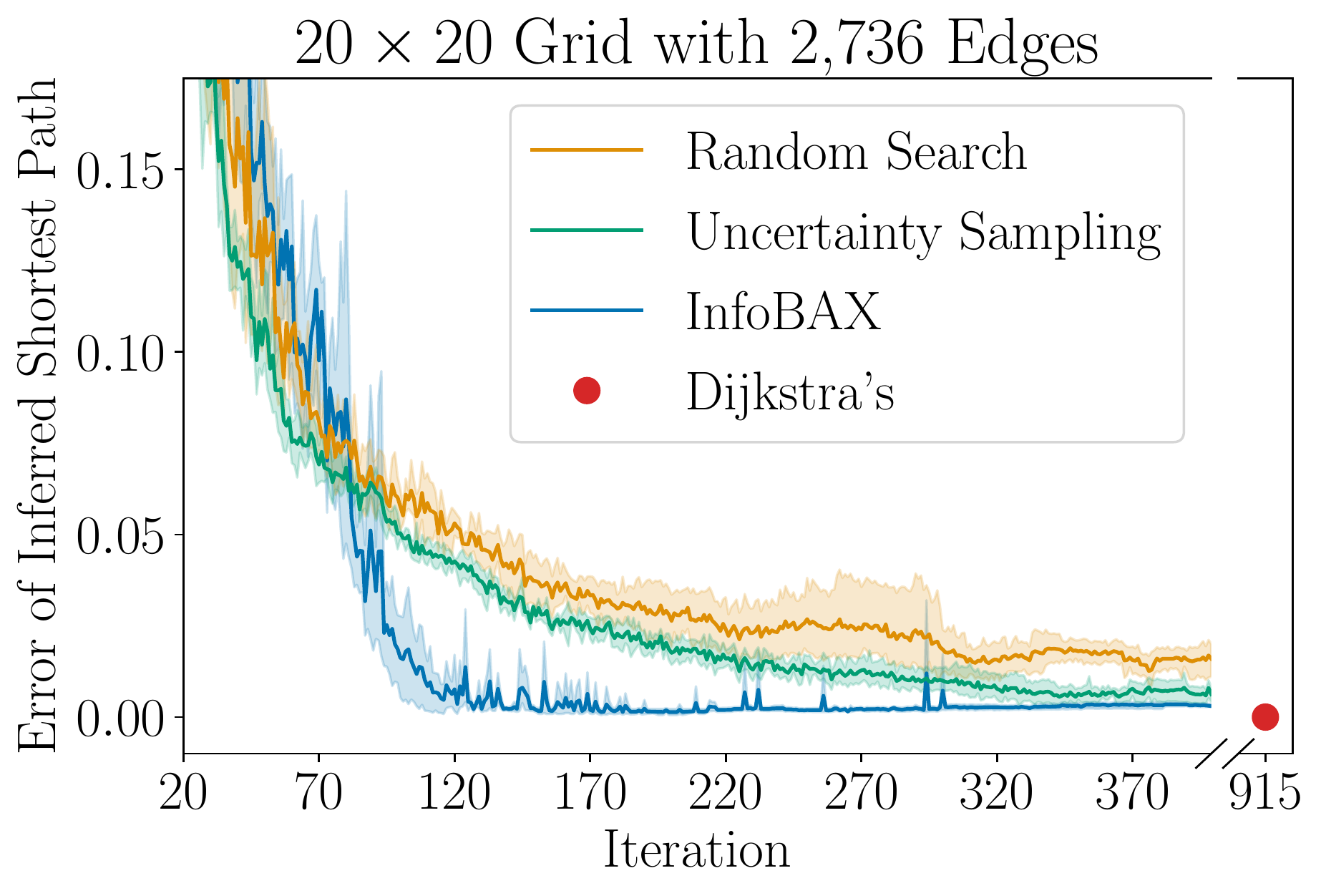}
\caption{\textbf{Estimating shortest paths in graphs:}
(Top row) A comparison of methods on grid-shaped graphs with edge weights give by
a black-box function, visualizing sample shortest paths (blue lines) produced by
running Dijkstra's algorithm on \(p(f \given \dataset_{T+1})\) for each method, given
a budget of $T=70$ queries. Black circles \circbox1{cblack} are queries, purple
squares \sqbox1{cpurple} are starting vertices, yellow squares \sqbox1{cyellow}
are destination vertices, pink circles \circbox1{cdeeppink} are the next queries,
ground truth shortest path is black dashed line.
(Middle row) Visualization of the 305 queries required by the full Dijkstra's algorithm.
(Bottom row) The error (sum of normalized polygonal areas between the inferred shortest
path and the ground truth shortest path) for each method vs. iteration, averaged over
5 trials.}
\label{fig:dijkstra_area}
\end{figure*}

\subsection{Estimating Shortest Paths in Graphs}
\label{sec:shortestpaths}

Finding the shortest path between two vertices in a graph is crucial in routing problems, such
as minimizing transportation costs, reducing latency when sending packets over the internet, and more.
Dijkstra's algorithm \citep{dijkstra1959note} provably recovers shortest paths by iteratively querying
edge costs as it searches a graph. However, in some applications, querying edge costs is expensive.
For example, in transportation networks, when edge costs represent the time required to traverse
unfamiliar terrain, it would be costly to survey each location in the order given by Dijkstra's algorithm.
Instead, we may try to survey a small set of locations that provide us with just enough information to
map out the shortest path through the terrain, avoiding the full evaluation cost of Dijkstra's.

As our first task, we use \infobax to infer the shortest path between two vertices in a graph
where the edge costs are represented by a black-box function. We use two synthetic graphs and
one real-world graph for our experiments. Our two synthetic graphs $(V,E)$ are grid-shaped with
$(|V|=10\times 10, |E|=684)$ and $(|V|=20\times 10, |E|=2736)$. We use the 2D Rosenbrock function
rescaled by $10^{-2}$ as the edge cost function for the synthetic graphs. Our real-world graph is
a cropped version of the \textit{California roads} network graph from \cite{li2005trip} and we use
the elevation of vertex midpoints from the Open-Elevation API as the edge cost function. Within
this graph, we seek to travel from Santa Cruz to Lake Tahoe for a nice change of outdoor activities.

We compare to baseline methods \randomsearch and \uncertaintysamplingnospace, which choose
$\datasett$ differently. \randomsearch forms $\datasett$ by random queries, while
\uncertaintysampling iteratively queries \(f\) at \(x\) that maximize the variance of
\(p(y_x \given \datasett)\). For \infobaxnospace, note that we can sample from
\(p(\algoutput \given \datasett)\) by executing algorithm \(\Acal\) on samples from
\(p(f \given \datasett)\). Since paths in our experiment consist of points in \(\Xcal\),
we use the aquisition function from Eq.~(\ref{eq:acqf_subexecpath}), choosing the points along
sampled shortests paths as our execution path subset. To evaluate the error between inferred shortest
path and the true shortest path in our planar graph, we use the polygonal area enclosed between the
inferred path and the true path. This geometrically captures deviations in the structure of the
inferred path from the true path. Notably, an inferred path recovers the ground truth if and only
if their enclosed area is zero. We normalize this error metric by the area of the overall graph.

Figure \ref{fig:dijkstra_area} (Top) shows this error metric between the inferred shortest paths
and the ground truth, averaged over inferred paths, with one standard error, in three experiments.
In all cases, \infobax recovers the ground truth shortest path using 5 to 547 times fewer queries
than would have been required to run Dijkstra's algorithm by itself. \infobax also outperforms the
baseline methods which fail to to recover the ground truth even with significantly more queries.

\begin{figure}[t]
\centering
\hspace{1mm}
\includegraphics[height=1.9in]{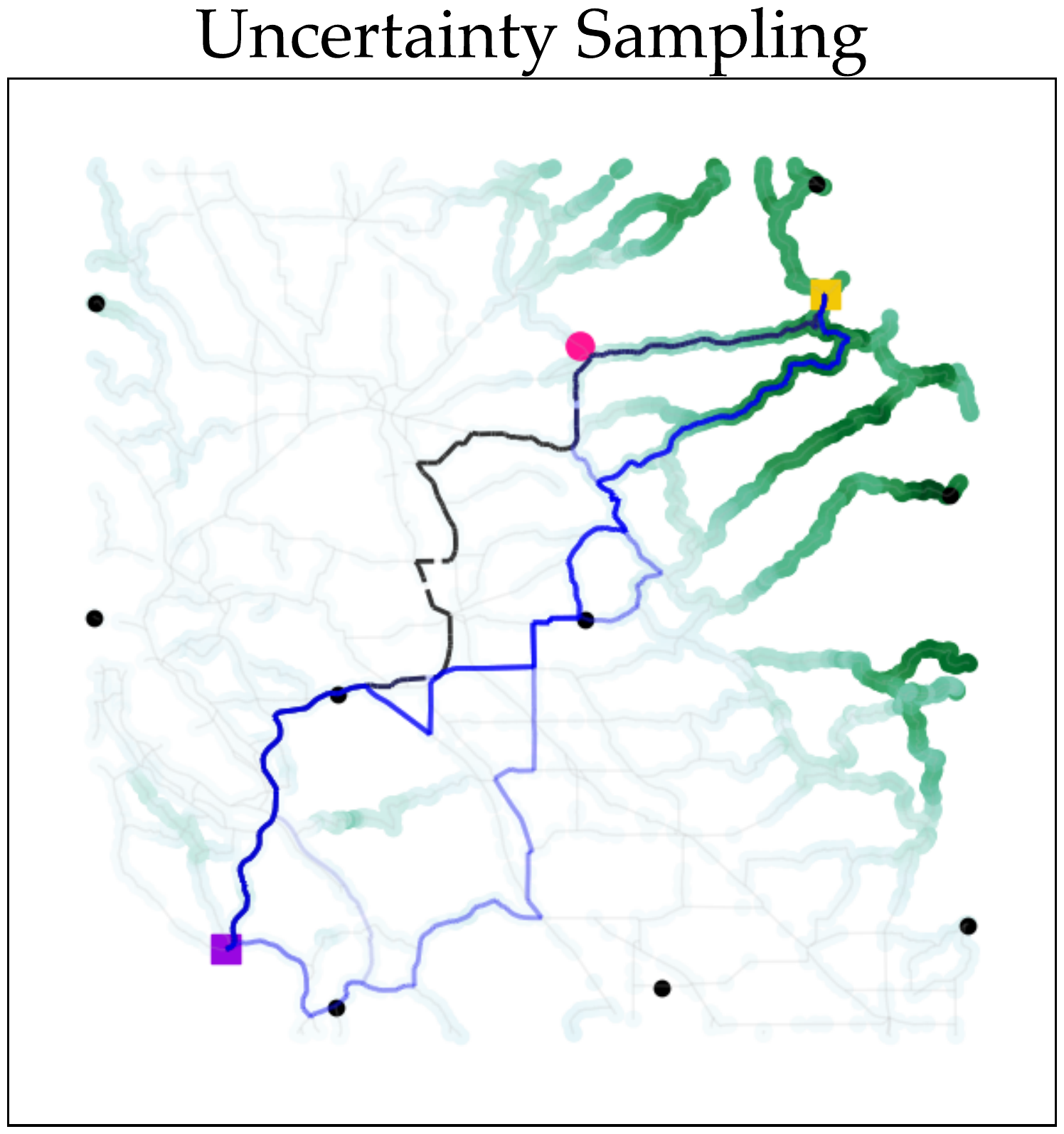}
\hspace{5mm}
\includegraphics[height=1.9in]{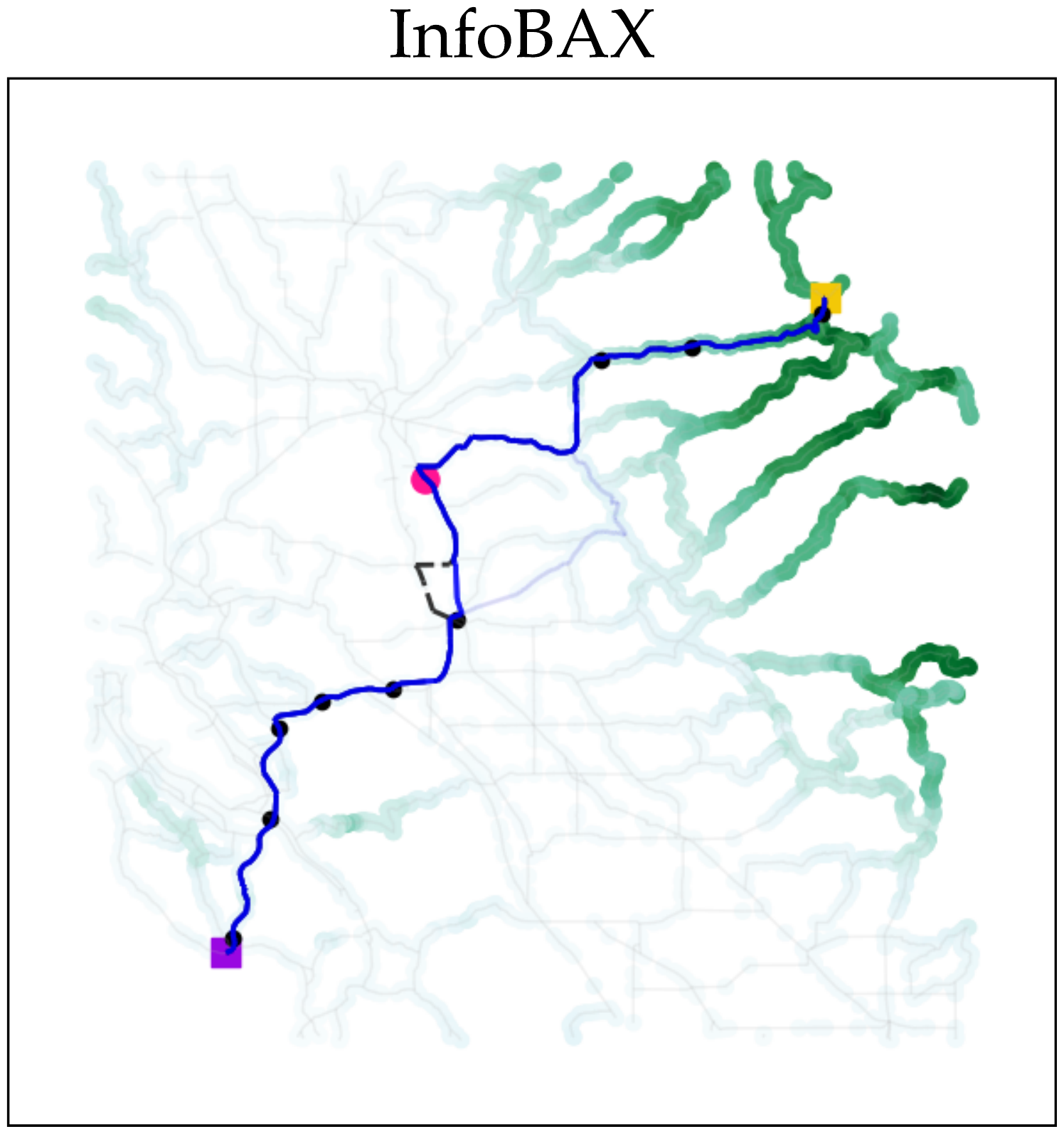}\\
\vspace{5mm}
$\vcenter{\hbox{\hspace{13mm}\includegraphics[height=1.9in]{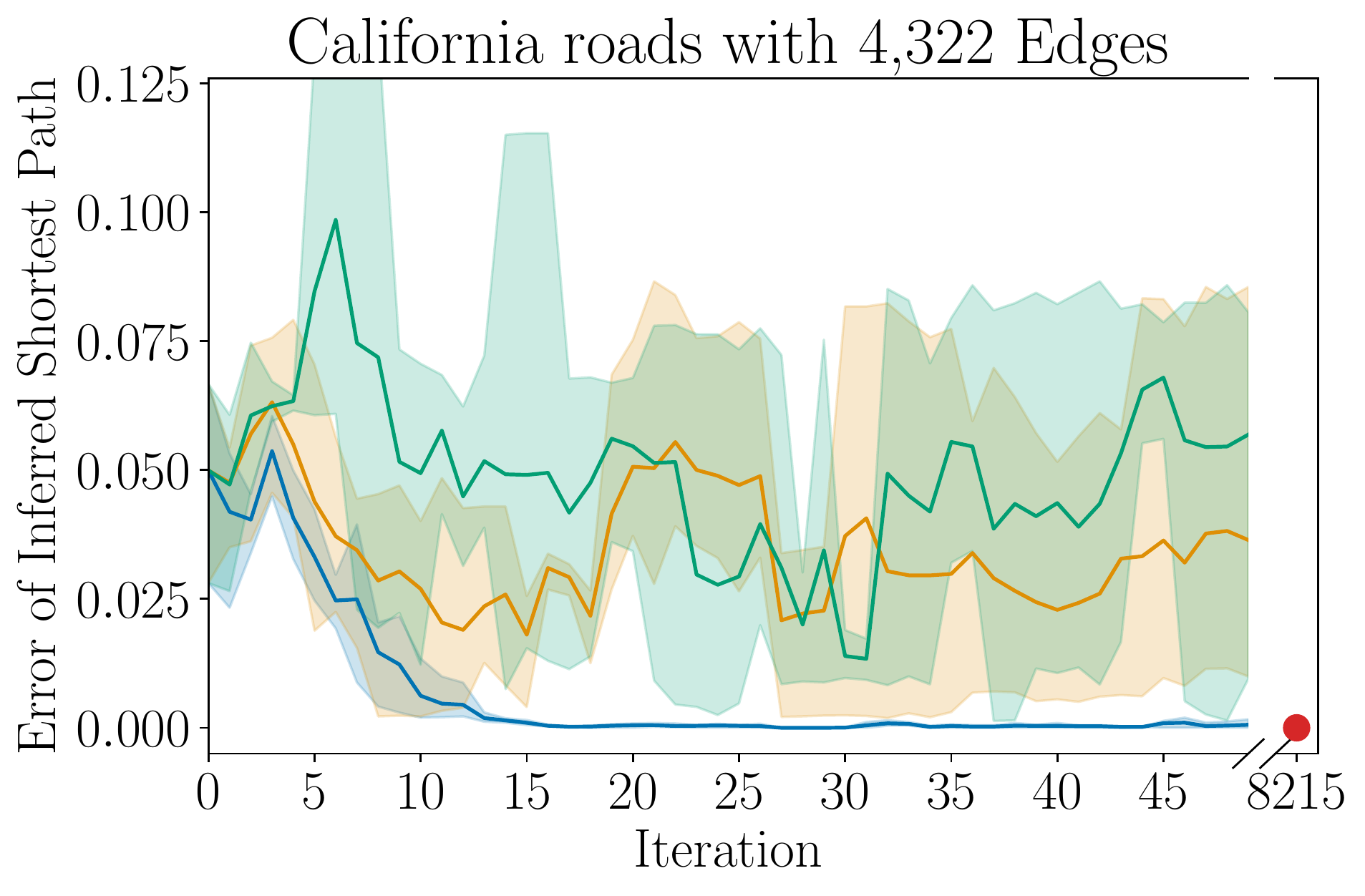}}}$
\hspace{-3mm}
$\vcenter{\hbox{\includegraphics[height=0.7in]{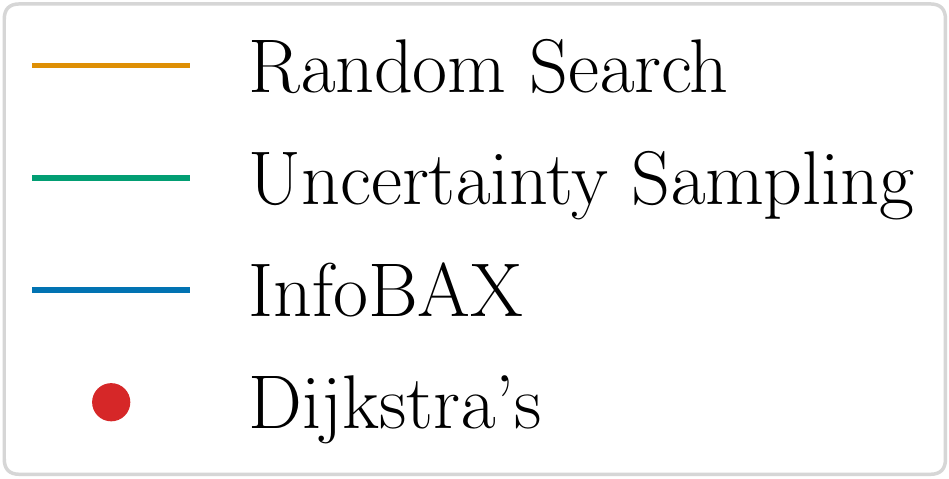}}}$
\caption{
\textbf{\textit{California roads} network:} (Top row) Visualization of inferences (blue lines)
of the true minimum-cost path (black dashed line) given by \uncertaintysampling (Left) and
\infobax (Right) after $T = 10$ queries.
(Bottom row) The error (sum of normalized polygonal areas between the inferred shortest path
and the ground truth shortest path) for each method vs. iteration, averaged over 5 trials.
}
\label{fig:dijkstra_ca}
\end{figure}

Figure \ref{fig:dijkstra_area} (Bottom) compares samples from the posterior distribution
\(p(\algoutput \given \datasett)\) given by \randomsearchnospace, \uncertaintysamplingnospace,
and \infobax queries. We see that \infobax spends its query budget around points that are most
informative about the shortest path, as expected. On the other hand,
\uncertaintysampling queries points that are informative about the overall function
$f$ but less informative about the property $\algoutput$. This behavior can also be
seen in Figure \ref{fig:dijkstra_ca} on the \textit{California roads} network.

Figure \ref{fig:dijkstra_area} (Top) shows this error metric between the inferred shortest paths and the ground truth, averaged over inferred paths, with one standard error, in three experiments.
In all cases, \infobax recovers the ground truth shortest path using 5 to 547 times fewer queries than would have been required to run Dijkstra's algorithm by itself. \infobax also outperforms the baseline methods which fail to to recover the ground truth even with significantly more queries.

Figure \ref{fig:dijkstra_area} (Bottom) compares samples from the posterior distribution \(p(\algoutput \given \datasett)\) by random queries, uncertainty-sampled queries, and \infobax queries. We see that \infobax spends its query budget around points that are most informative about the shortest path, as expected. On the other hand, \textsc{UncertaintySampling} queries points that are informative about the overall function $f$ but less informative about the property $\algoutput$. This behavior can also be seen in Figure \ref{fig:dijkstra_ca} on the \textit{California roads} network.

\begin{figure*}[t]
\centering
\includegraphics[height=2.0in]{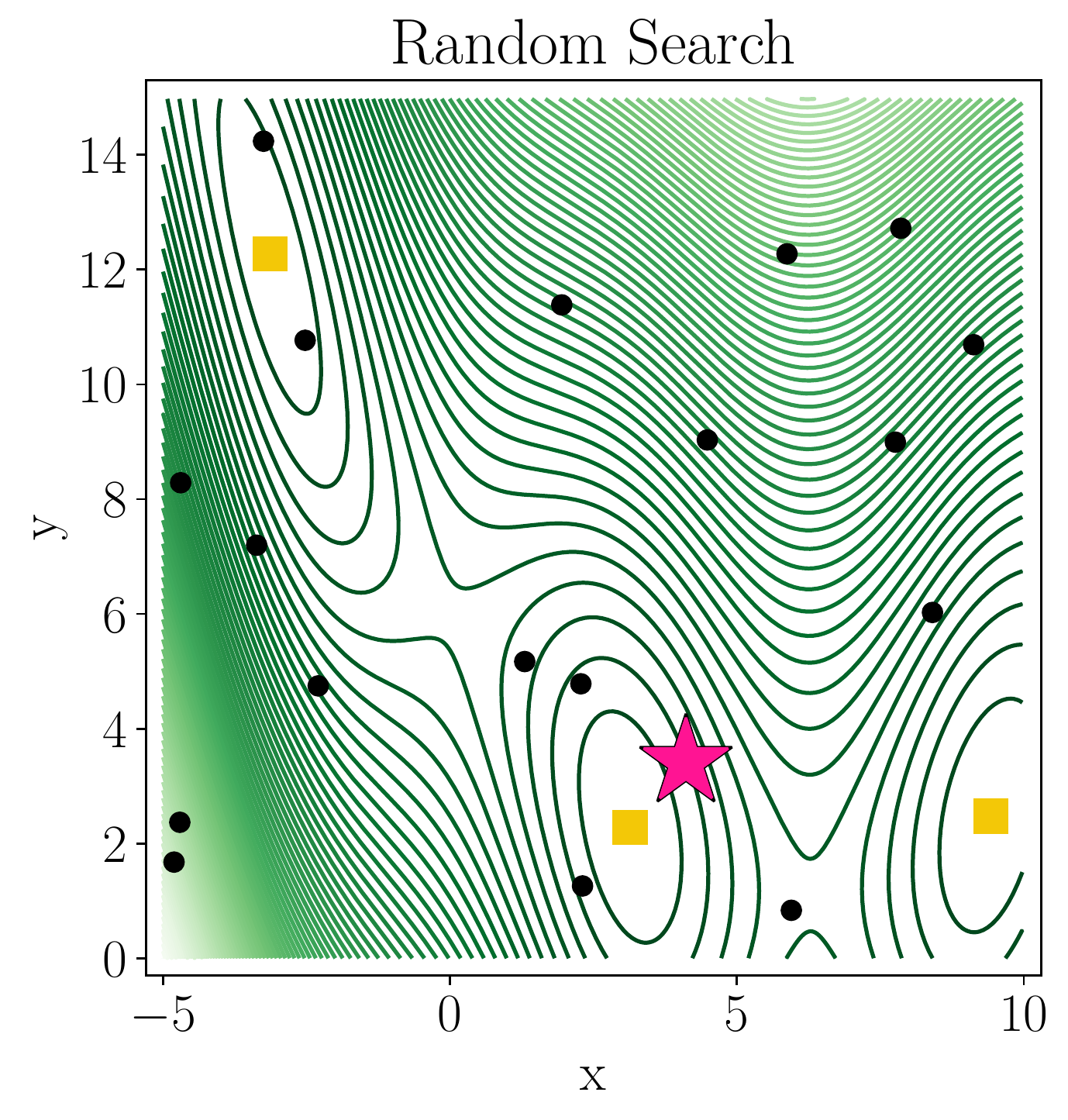}
\includegraphics[height=2.0in]{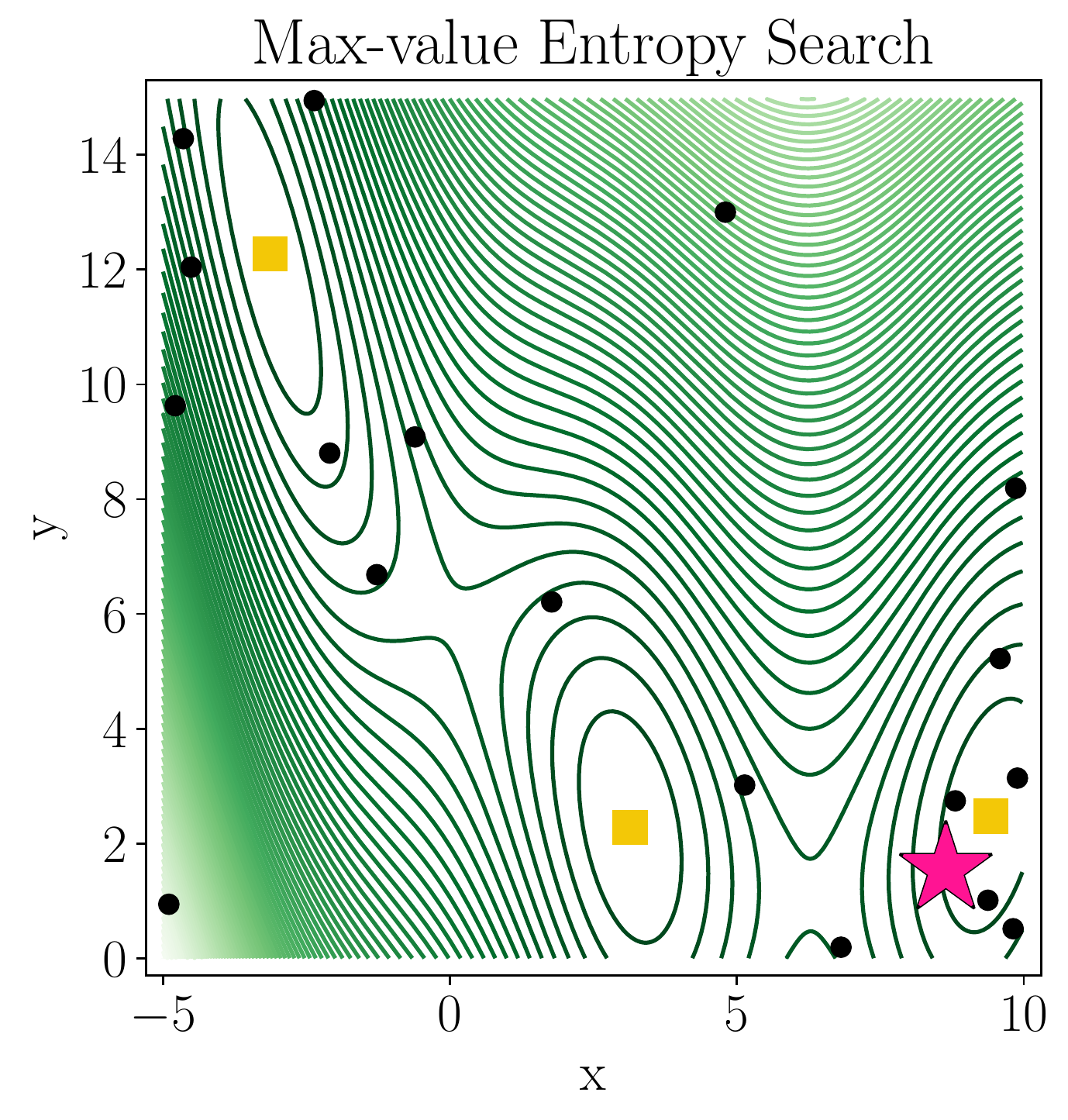}
\includegraphics[height=2.0in]{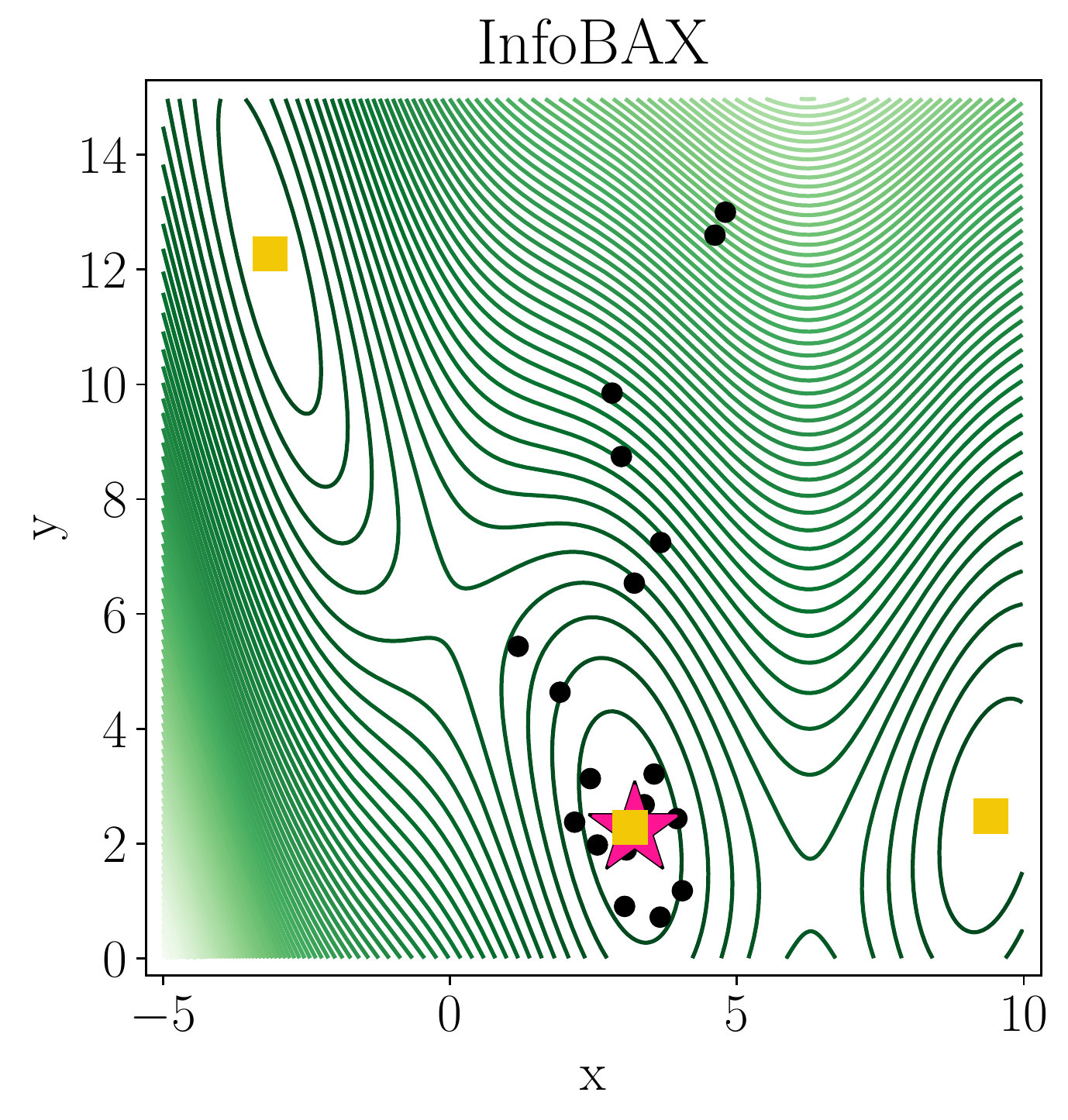}
\includegraphics[height=2.0in]{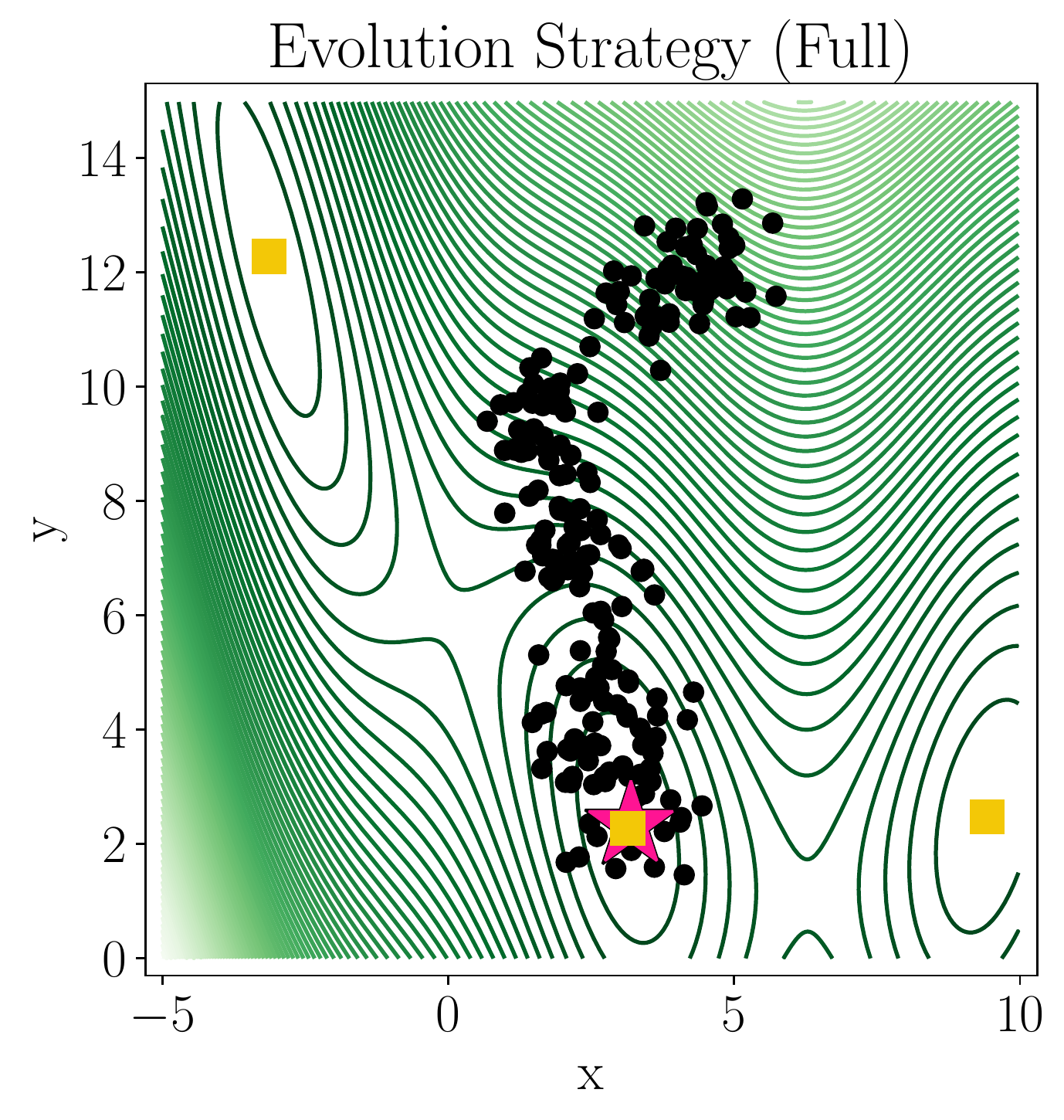}\\
\includegraphics[height=1.9in]{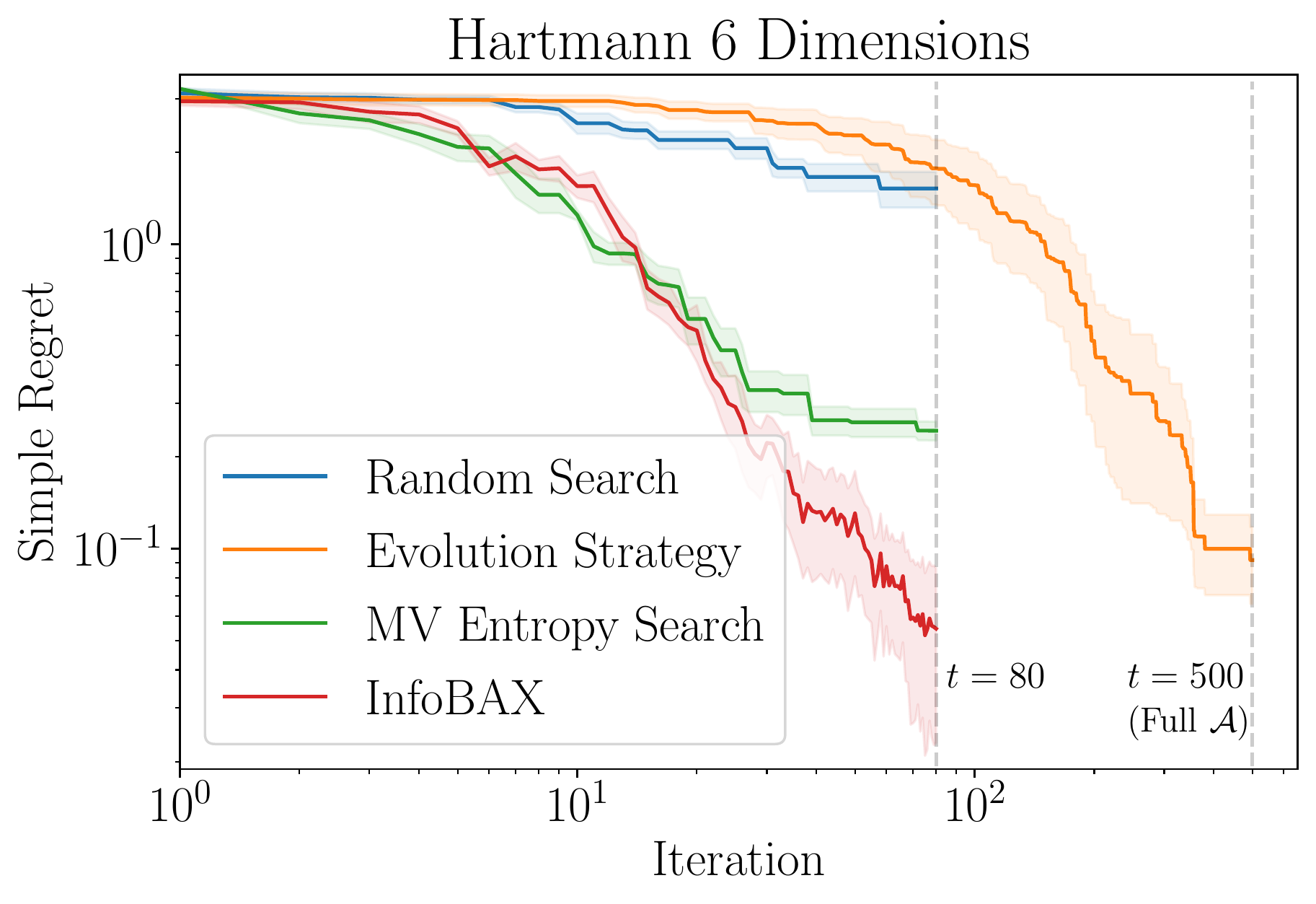}
\hspace{2mm}
\includegraphics[height=1.9in]{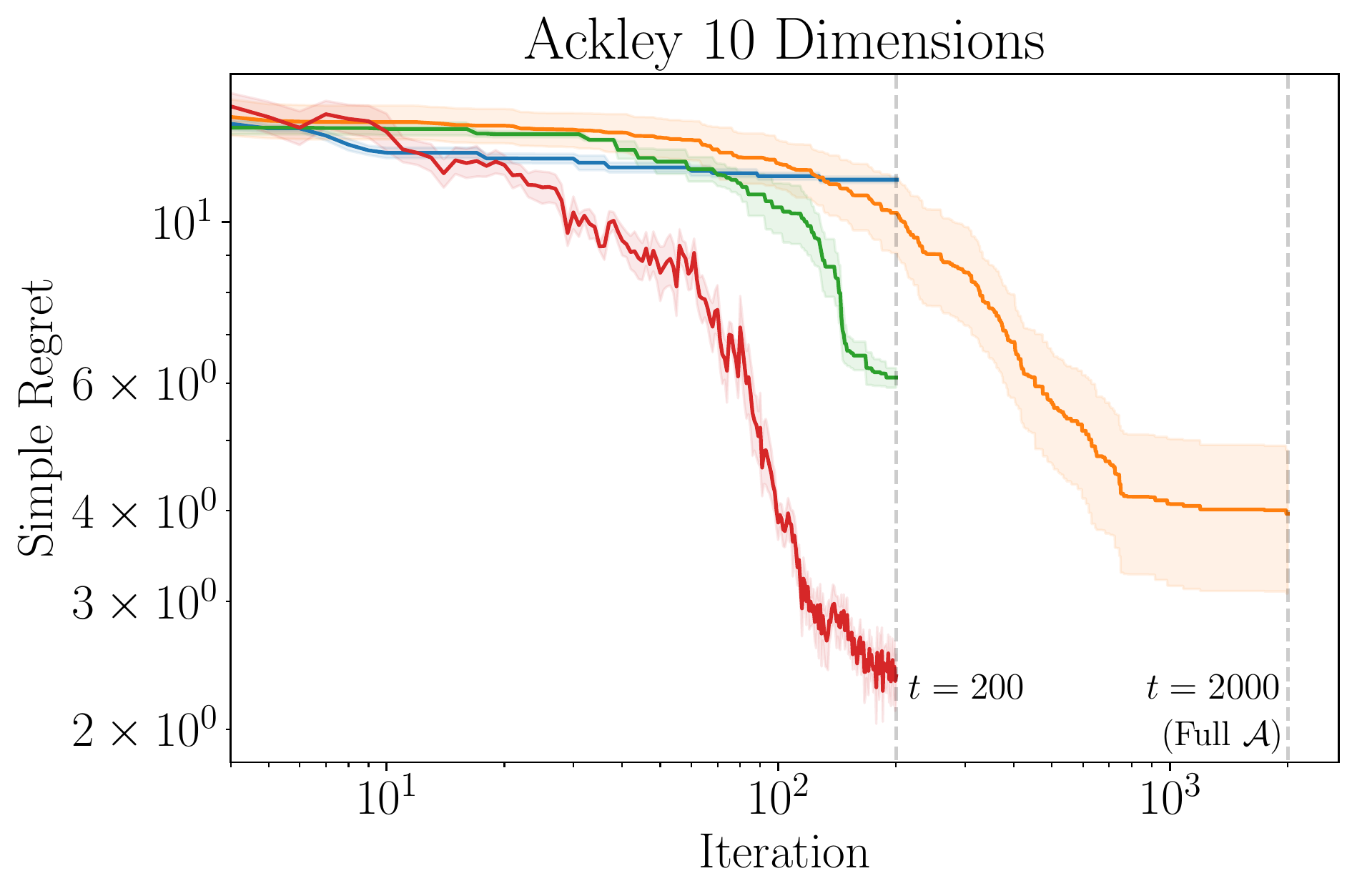}
\caption{
\textbf{Bayesian local optimization:} (Top row) Visualization of function queries and
estimated optima for each method, given a budget of $T=18$ queries. Black
circles~\circbox1{cblack} are function queries, pink stars~$\color{magenta} \bigstar$
are estimated optima, and yellow squares \sqbox1{cyellow} are the true optima.
(Middle row) Queries made by the full \textsc{EvolutionStrategy} algorithm ($T=208$)
without \infobaxnospace.
(Bottom row) The difference between the value $f(\hat{x})$ at an
estimated optimum $\hat{x}$ and the true optimal value $f(x^*)$,
vs. iteration, on two benchmark tasks.
}
\label{fig:localopt}
\vspace{-3mm}
\end{figure*}

\subsection{Bayesian Local Optimization}

Bayesian optimization is a popular method for probabilistic model-based
\textit{global optimization} \citep{shahriari2015taking, Frazier2018-bc},
that aims to determine global optima of a black-box $f$ in a query-efficient manner.
There also exist many \textit{local optimization} algorithms, such as evolution
strategies \citep{back1996evolutionary}, the Nelder-Mead algorithm \citep{nelder1965simplex},
COBYLA \citep{powell1994direct}, and finite-difference gradient descent
procedures \citep{richardson1911ix, spall1992multivariate}, for optimizing a black-box $f$.
In certain settings these algorithms have shown very strong performance, such as when $\Xcal$
is high-dimensional, and when function evaluations are cheap and many queries of $f$ can be
made \citep{rios2013derivative}. This is potentially because they do not explore as broadly to
explicitly try and find a global optima and instead greedily optimize to nearby local optima,
or potentially due to other aspects of their updates and how they traverse the space.
Regardless, under the right conditions, these algorithms can often be applied to great effect.

However, when function evaluations are expensive, local optimization methods can suffer:
these algorithms are often query-\textit{inefficient}, and may perform a large number of
similar evaluations, which hurts performance significantly.
Here, Bayesian optimization methods tend to show better
performance \citep{eriksson2019scalable, letham2020re}.
Furthermore, these local methods may not be suited for settings with certain
function noise which can be handled more easily in Bayesian optimization via a custom model.

Ideally, we would like the best of both worlds: a procedure that incorporates the
model-induced query-efficiency of Bayesian optimization, and also takes advantage
of the greedy optimization strategies provided by various local optimization
algorithms (which are effective if only they were applied directly to a cheap,
noiseless $f$).

We therefore propose running \infobax on a local optimization algorithm
$\Acal$, which yields a variant of Bayesian optimization that we refer to
as \textit{Bayesian local optimization}.
Here, the main idea is that we approach Bayesian optimization as the task of
inferring the output $\algoutput$ of a local optimization
algorithm run on $f$---rather than estimating a global optima of $f$---using
as few queries as possible.

We demonstrate this procedure by implementing $\Acal$ as an evolution strategy,
where a population of vectors are randomly mutated and pruned based on their objective values
(details given in Section~\ref{sec:experimentsappendix}).
We compare \infobax against this \textsc{EvolutionStrategy},
and also against both \textsc{RandomSearch} and
\textsc{MaxValueEntropySearch} \citep{Wang2017-fb}, which is a popular
information-based Bayesian optimization method that aims to efficiently
infer global optima of $f$.

We show results on black-box function optimization benchmark tasks.
Figure~\ref{fig:localopt} (Top) compares evaluations chosen by the four methods,
where the first three plots show results at $T = 18$ iterations,
while the fourth plot shows the full \textsc{EvolutionStrategy} ($T=208$).
\infobax is able to estimate $\algoutput$ (pink star) using only a fraction of the queries.

Figure~\ref{fig:localopt} (Bottom) shows the difference between the value of
$f(\hat{x})$ at an estimated optimum $\hat{x}$ versus the true optimal value $f(x^*)$
(over five trials, showing one standard error), on two benchmark functions with domains
$\Xcal$ in six and ten dimensions. 
In both cases, \infobax outperforms the baselines and is able to match the eventual performance
of the \textsc{EvolutionStrategy} using 8 to 20 times fewer function evaluations.

\subsection{Top-$k$ estimation}
\label{sec:experiments-topk}

We show additional experimental results on  the problem of top-$k$ estimation.
In Section~\ref{sec:methodoverview}, we describe the task of top-$k$ estimation,
which we summarize here as follows.
Suppose we have a finite collection of elements $X \subseteq \Xcal$, where each
$x \in X$ has an unknown value $f_x$.
There are many applications where we care about estimating the
\textit{top-$k$ elements of $X$} with the highest values, denoted $K^* \subseteq X$.
Given a budget $T$, our goal will be to choose the best
$T$ inputs $x_1, \ldots, x_T$ to query, in order to infer $K^*$.
For full generality, assume that we can evaluate any $x_t \in \Xcal$,
so we are not restricted to evaluating only inputs in $X$.
This problem can be viewed as a type of active search,
which extends optimization to estimating the top-$k$, rather than top-1, 
element in a discrete set. It also has relations to level set 
estimation, where the goal is to estimate all elements $x \in X$ with 
a value $f_x$ above some threshold $C$.

\begin{figure*}[t]
    \centering
    \includegraphics[height=1.85in]{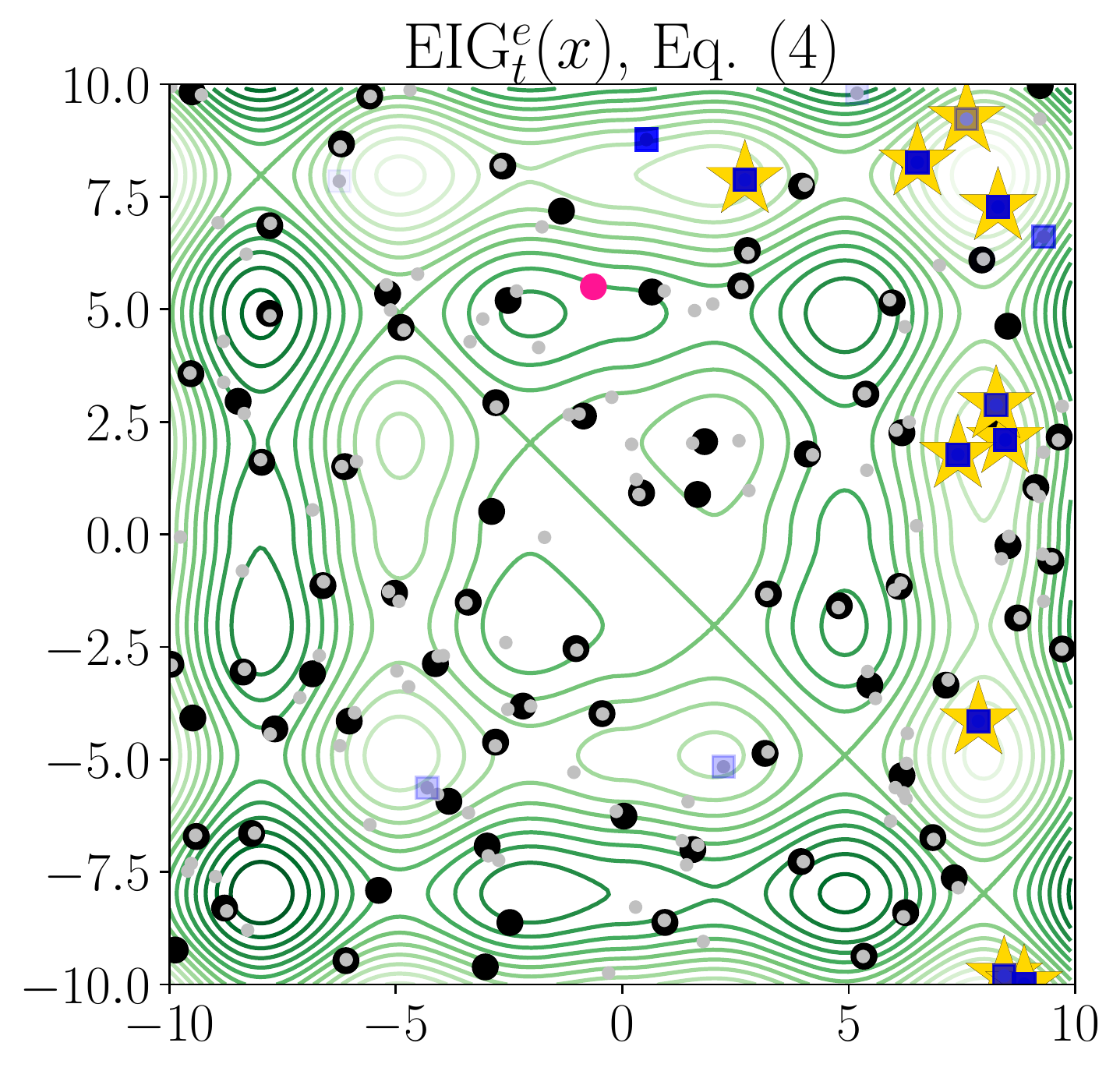}
    \includegraphics[height=1.85in]{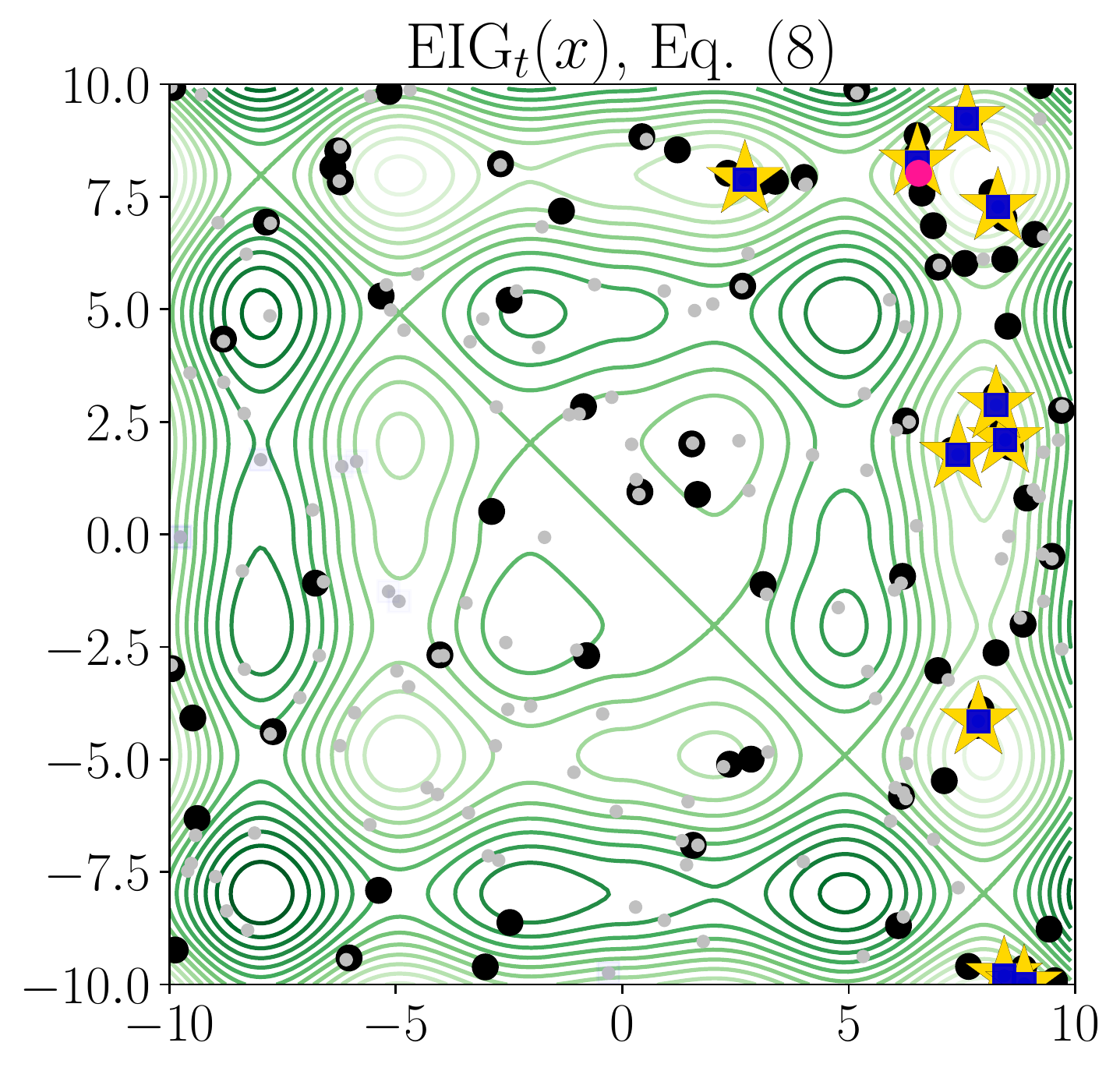}
    \includegraphics[height=1.85in]{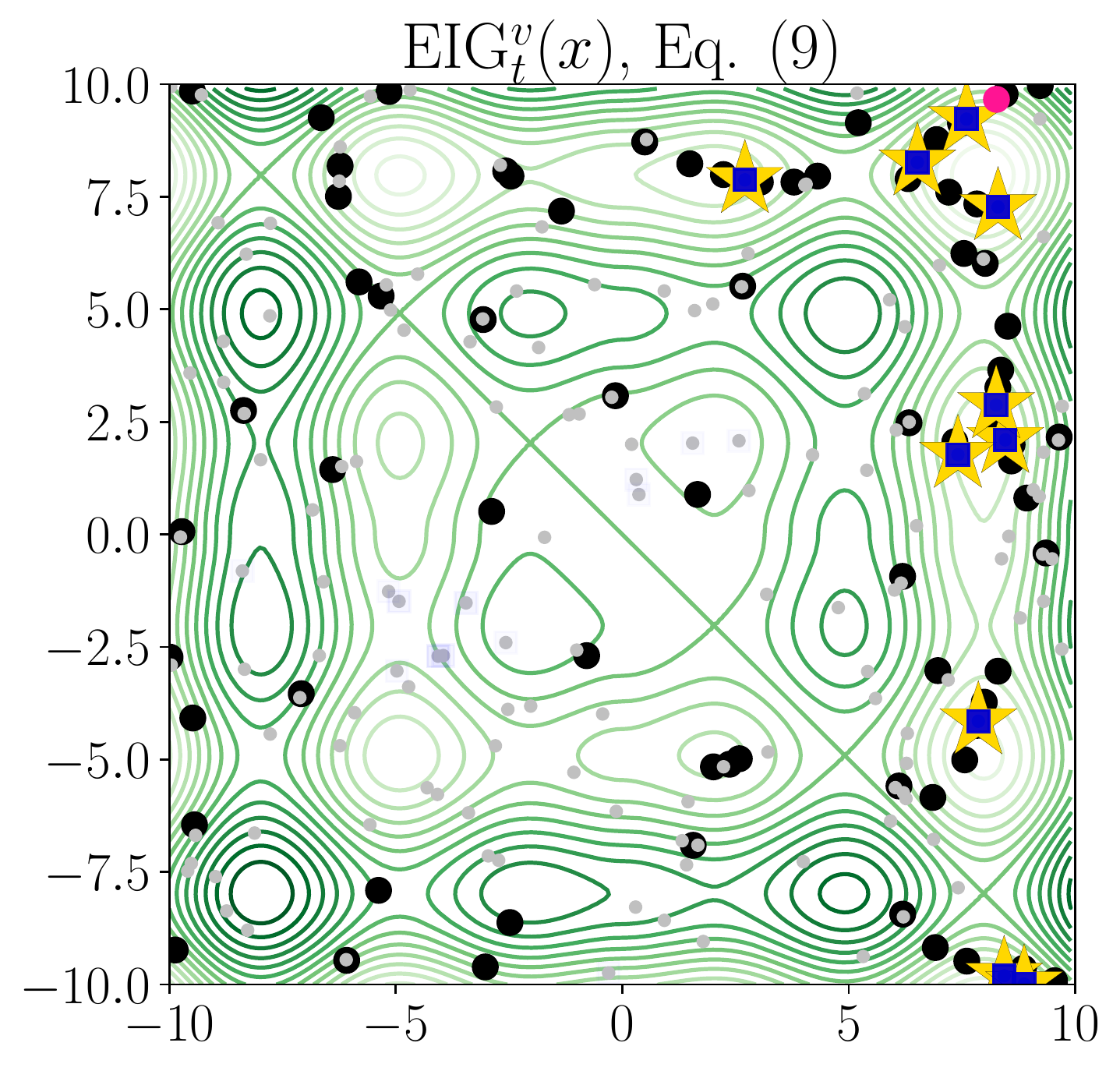}\\
    \includegraphics[height=1.85in]{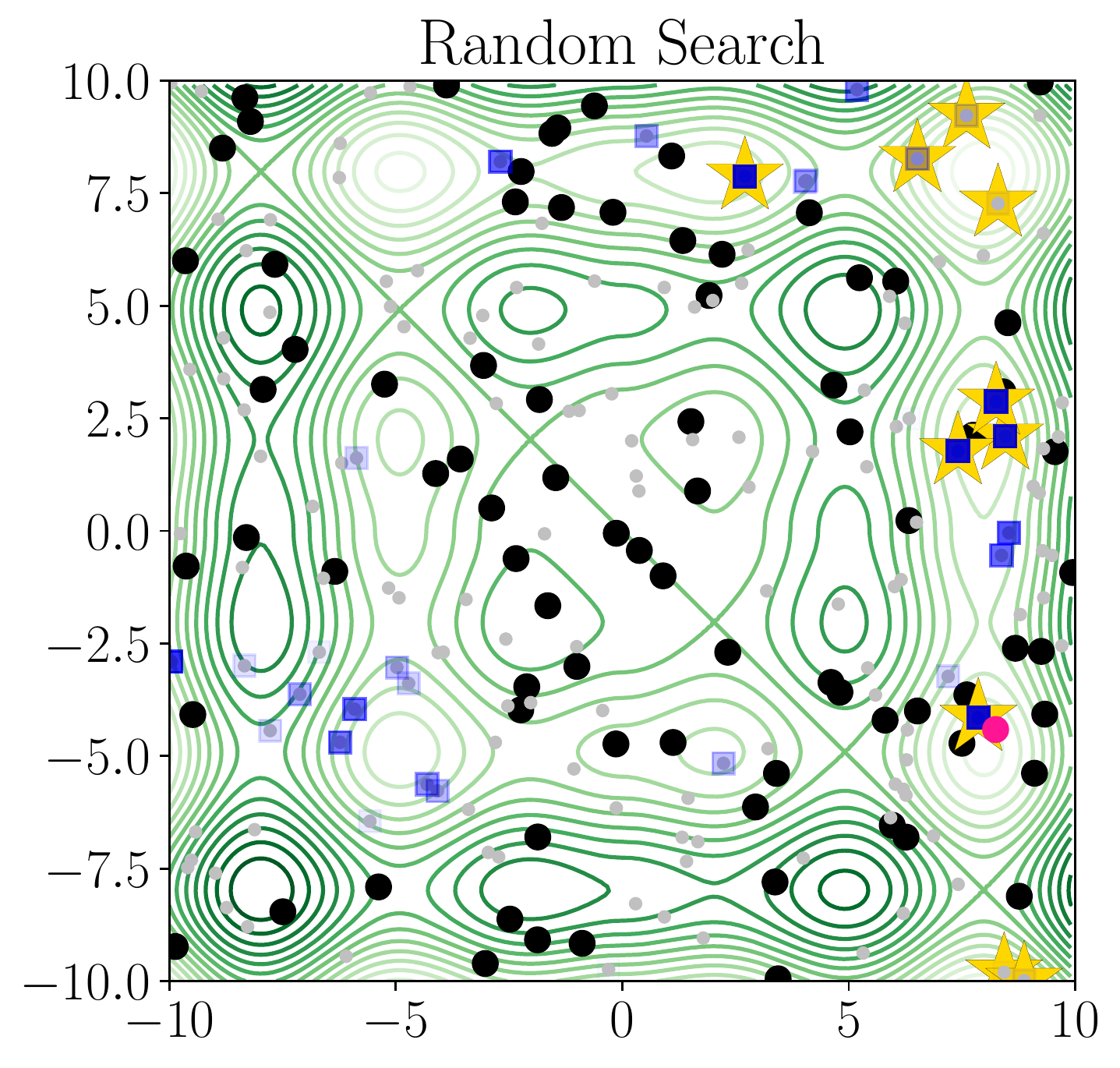}
    \includegraphics[height=1.85in]{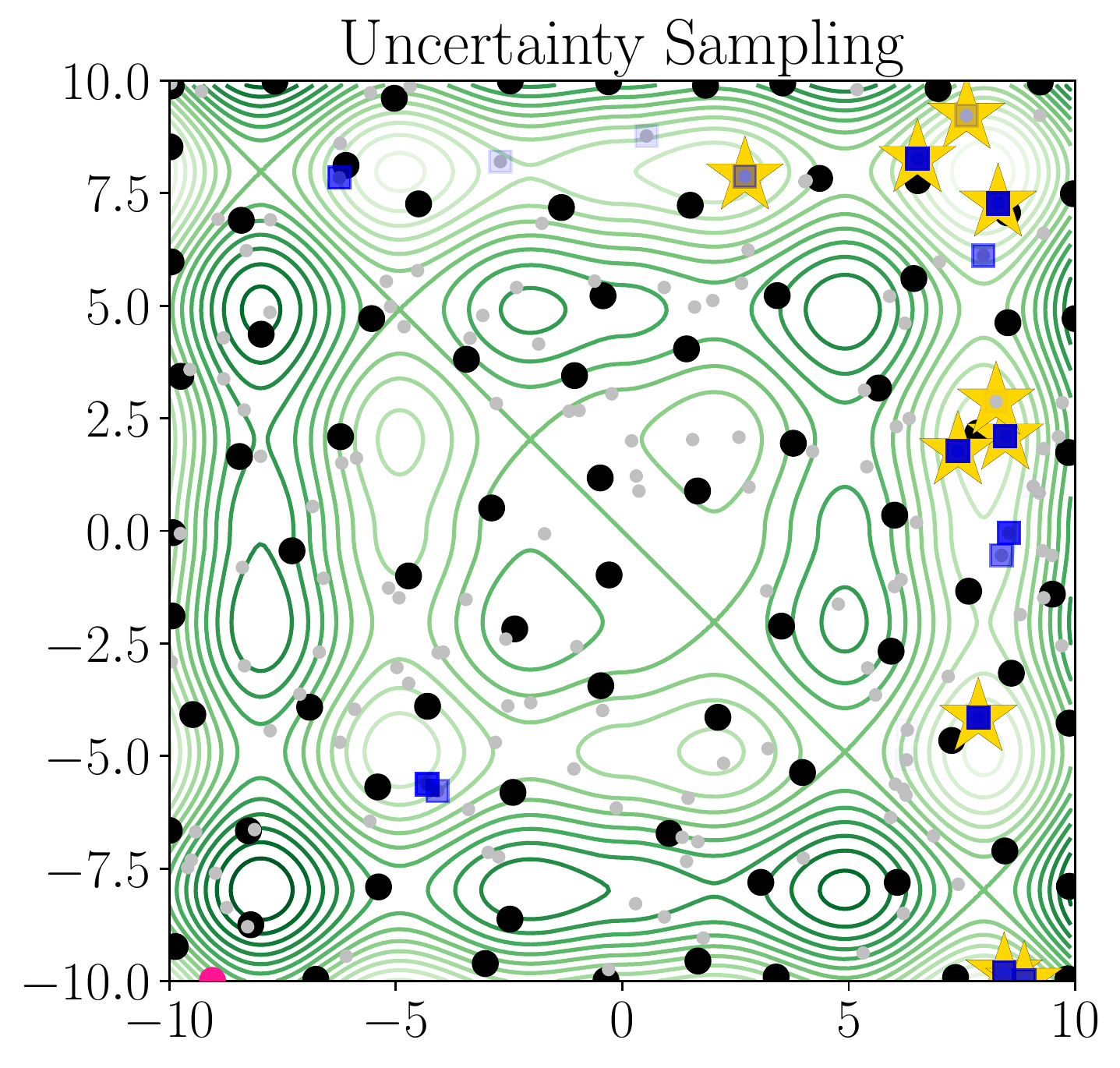}\\
    \includegraphics[height=1.9in]{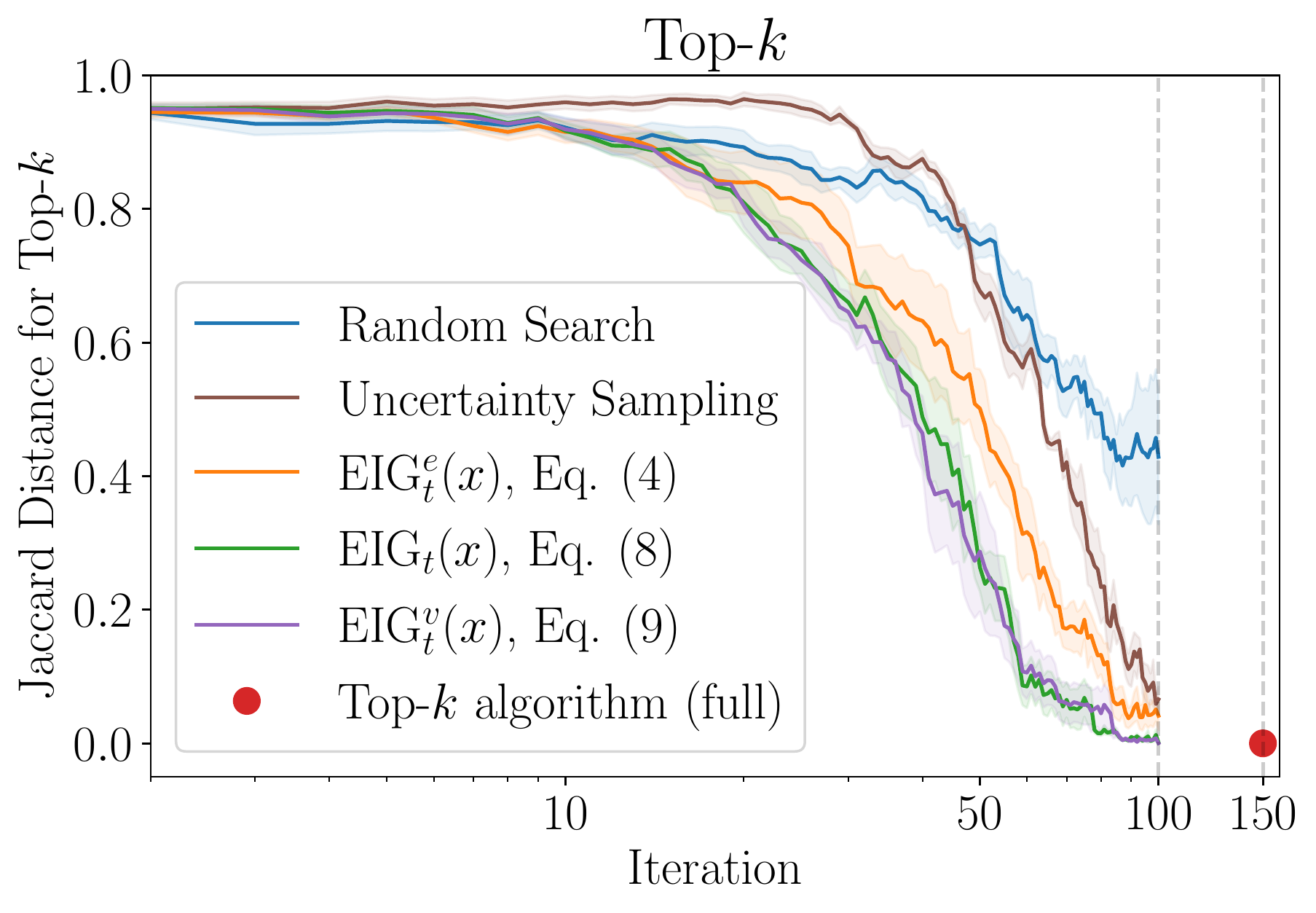}
    \caption{
    \textbf{Top-$k$ estimation results:}
    (Top two rows) Visualization of methods, where light grey dots $\color{gray} \bullet$
    are the 150 elements $X \subset \Xcal$ which comprise the execution path $\exepath$,
    black circles \circbox1{cblack} are function evaluations, gold stars
    $\color{goldenpoppy} \bigstar$ are the true top $k=10$ elements with highest value $f_x$, 
    pink circles \circbox1{cdeeppink} are the next evaluation chosen, and blue
    squares \sqbox1{cblue} are posterior samples of the output (top-$k$ elements).
    For each method, $T=70$ evaluations are shown.
    (Bottom row) The Jaccard distance averaged over 5 trials vs. iteration,
    with error bars showing one standard error.
    }
    \label{fig:topkresults}
    \vspace{-3mm}
\end{figure*}

To run \infobax for this problem, we make use of the following
\textit{top-$k$ algorithm} $\Acal$:
evaluate  $f_x$ for each $x \in X$, sort $X$ in decreasing order, and return the 
first $k$ elements. This algorithm makes exactly $|X|$ evaluations of $f$.
In Figure~\ref{fig:topk}, we illustate the top-$k$ algorithm $\Acal$, 
as well as the three acquisition functions in Eqs.~(\ref{eq:acqf_execpath}),
(\ref{eq:acqf_output_expect}), and (\ref{eq:acqf_subexecpath}).

We carry out a top-$k$ estimation experiment on a two dimensional domain
$\Xcal \in \Rbb^2$, where for each $x \in \Xcal$, $-10 < x_1 < 10$,
and $-10 < x_2 < 10$. From this domain, we draw a set of 150 elements
$X$ uniformly at random, and choose to estimate the top $k = 10$ elements.
For this experiment, we use the skewed sinusoidal function
$g: \Xcal \rightarrow \Rbb$, defined as $g(x) = \sum_{i=1}^d 2 |x_i| \sin(x_i)$,
which has a multimodal landscape.

Our goal is then to infer $K^* \subseteq X$, 
the top-$k$ elements of $X$, using as few queries of $f$ as possible.
We compare the performance of our three \infobax acquisition 
functions ($\acqfnt^e(x)$, $\acqfnt(x)$, and $\acqfnt^v(x)$), along
with \textsc{RandomSearch} and \textsc{UncertaintySampling}
(both decribed in Section~\ref{sec:shortestpaths}), as well as the
full top-$k$ algorithm that scans through each point in $X$.
For \infobax methods, we draw 100 samples of $\samp{f} \sim p(f \given \datasett)$ 
and run the top-$k$ algorithm $\Acal$ on these, in order to produce the 
execution path samples $\exepathsamp$, or algorithm output samples $\algoutputsamp$.

In Figure~\ref{fig:topkresults} (Bottom) we show these results, plotting the Jaccard
distance for each method at each iteration, which, for a given estimate $\hat{K}$ of
the top-$k$ elements of $X$, is defined as
\begin{align}
    \text{Jaccard distance}(\hat K, K^*)  
    = 1 - \text{Jaccard index}(\hat K, K^*) 
    = 1 - \frac{|\hat{K} \cap K^*|}{|\hat{K} \cup K^*|}.
\end{align}
For each method, we average this metric over five trials and show one standard error.
The \infobax acquisition functions $\acqfnt^v(x)$ and $\acqfnt(x)$ accurately infer
the true top-$k$ set in the fewest iterations (using roughly 2 times fewer function
evaluations than the full top-$k$ algorithm), followed by \infobax using $\acqfnt^e(x)$, 
\textsc{UncertaintySampling}, and finally \textsc{RandomSearch}.

In Figure~\ref{fig:topkresults} (Top) we show the set of function evaluations
and posterior samples of the inferred top-$k$ sets $\hat{K}$ for each method.
We see that \infobaxnospace, using $\acqfnt^v(x)$ and $\acqfnt(x)$, is able to determine
and spend its query budget around the true top-$k$ elements $K^*$
(denoted by gold stars). Note also that \infobax using $\acqfnt^e(x)$
focuses its query budget on the execution path $\exepath$ (or, 
equivalently, the set $X$), shown as light grey dots,
while \textsc{UncertaintySampling} spends its budget on points that are
informative about the full function $f$, as opposed to the execution path
$\exepath$ or top-$k$ property $\algoutput$.
\vspace{-1mm}
\section{Conclusion}
\label{sec:conclusion}

The BAX framework unifies problems in disparate domains that seek to estimate properties of
black-box functions given limited function evaluations. For a property-computing algorithm
\(\Acal\), our proposed method, \infobaxnospace, is able to make targeted queries that can
reduce function evaluations by up to hundreds of times \emph{without modifying \(\Acal\) to
respect the budget constraint}. However, \infobax also has its limitations. For example,
it may be difficult to find an appropriate model \(p(f)\), in certain settings.
Nevertheless, when we have an accurate function prior, we can
dramatically offload the cost of function evaluations to the cost of parallelizable computations.
In the future, we hope this branch of methods could potentially aid in custom optimization
tasks in the sciences \citep{char2019offline},
interactive human-in-the-loop methods \citep{boecking2020interactive},
and fields such as drug and materials discovery,
where function evaluations may be highly expensive or time consuming.

\bibliography{main}

\begin{thebibliography}{56}
\providecommand{\natexlab}[1]{#1}
\providecommand{\url}[1]{\texttt{#1}}
\expandafter\ifx\csname urlstyle\endcsname\relax
  \providecommand{\doi}[1]{doi: #1}\else
  \providecommand{\doi}{doi: \begingroup \urlstyle{rm}\Url}\fi

\bibitem[Adafre and de~Rijke(2005)]{adafre2005discovering}
Sisay~Fissaha Adafre and Maarten de~Rijke.
\newblock Discovering missing links in wikipedia.
\newblock In \emph{Proceedings of the 3rd international workshop on Link
  discovery}, pages 90--97, 2005.

\bibitem[Back(1996)]{back1996evolutionary}
Thomas Back.
\newblock \emph{Evolutionary algorithms in theory and practice: evolution
  strategies, evolutionary programming, genetic algorithms}.
\newblock Oxford university press, 1996.

\bibitem[Backstrom and Leskovec(2011)]{backstrom2011supervised}
Lars Backstrom and Jure Leskovec.
\newblock Supervised random walks: predicting and recommending links in social
  networks.
\newblock In \emph{Proceedings of the fourth ACM international conference on
  Web search and data mining}, pages 635--644, 2011.

\bibitem[Bast et~al.(2016)Bast, Delling, Goldberg, M{\"u}ller-Hannemann, Pajor,
  Sanders, Wagner, and Werneck]{bast2016route}
Hannah Bast, Daniel Delling, Andrew Goldberg, Matthias M{\"u}ller-Hannemann,
  Thomas Pajor, Peter Sanders, Dorothea Wagner, and Renato~F Werneck.
\newblock Route planning in transportation networks.
\newblock In \emph{Algorithm engineering}, pages 19--80. Springer, 2016.

\bibitem[Beaumont(2010)]{beaumont2010approximate}
Mark~A Beaumont.
\newblock Approximate bayesian computation in evolution and ecology.
\newblock \emph{Annual review of ecology, evolution, and systematics},
  41:\penalty0 379--406, 2010.

\bibitem[Beaumont et~al.(2002)Beaumont, Zhang, and
  Balding]{beaumont2002approximate}
Mark~A Beaumont, Wenyang Zhang, and David~J Balding.
\newblock Approximate bayesian computation in population genetics.
\newblock \emph{Genetics}, 162\penalty0 (4):\penalty0 2025--2035, 2002.

\bibitem[Bect et~al.(2012)Bect, Ginsbourger, Li, Picheny, and
  Vazquez]{bect2012sequential}
Julien Bect, David Ginsbourger, Ling Li, Victor Picheny, and Emmanuel Vazquez.
\newblock Sequential design of computer experiments for the estimation of a
  probability of failure.
\newblock \emph{Statistics and Computing}, 22\penalty0 (3):\penalty0 773--793,
  2012.

\bibitem[Boecking et~al.(2020)Boecking, Neiswanger, Xing, and
  Dubrawski]{boecking2020interactive}
Benedikt Boecking, Willie Neiswanger, Eric Xing, and Artur Dubrawski.
\newblock Interactive weak supervision: Learning useful heuristics for data
  labeling.
\newblock \emph{arXiv preprint arXiv:2012.06046}, 2020.

\bibitem[Caselton and Zidek(1984)]{caselton1984optimal}
William~F Caselton and James~V Zidek.
\newblock Optimal monitoring network designs.
\newblock \emph{Statistics \& Probability Letters}, 2\penalty0 (4):\penalty0
  223--227, 1984.

\bibitem[Chaloner and Verdinelli(1995)]{Chaloner1995-op}
Kathryn Chaloner and Isabella Verdinelli.
\newblock Bayesian experimental design: A review.
\newblock \emph{Stat. Sci.}, 10\penalty0 (3):\penalty0 273--304, 1995.

\bibitem[Char et~al.(2019)Char, Chung, Neiswanger, Kandasamy, Nelson, Boyer,
  Kolemen, and Schneider]{char2019offline}
Ian Char, Youngseog Chung, Willie Neiswanger, Kirthevasan Kandasamy, Andrew~O
  Nelson, Mark Boyer, Egemen Kolemen, and Jeff Schneider.
\newblock Offline contextual bayesian optimization.
\newblock \emph{Advances in Neural Information Processing Systems},
  32:\penalty0 4627--4638, 2019.

\bibitem[Chevalier et~al.(2014)Chevalier, Bect, Ginsbourger, Vazquez, Picheny,
  and Richet]{chevalier2014fast}
Cl{\'e}ment Chevalier, Julien Bect, David Ginsbourger, Emmanuel Vazquez, Victor
  Picheny, and Yann Richet.
\newblock Fast parallel kriging-based stepwise uncertainty reduction with
  application to the identification of an excursion set.
\newblock \emph{Technometrics}, 56\penalty0 (4):\penalty0 455--465, 2014.

\bibitem[Csill{\'e}ry et~al.(2010)Csill{\'e}ry, Blum, Gaggiotti, and
  Fran{\c{c}}ois]{csillery2010approximate}
Katalin Csill{\'e}ry, Michael~GB Blum, Oscar~E Gaggiotti, and Olivier
  Fran{\c{c}}ois.
\newblock Approximate bayesian computation (abc) in practice.
\newblock \emph{Trends in ecology \& evolution}, 25\penalty0 (7):\penalty0
  410--418, 2010.

\bibitem[Davis and Rabinowitz(2007)]{davis2007methods}
Philip~J Davis and Philip Rabinowitz.
\newblock \emph{Methods of numerical integration}.
\newblock Courier Corporation, 2007.

\bibitem[Dijkstra et~al.(1959)]{dijkstra1959note}
Edsger~W Dijkstra et~al.
\newblock A note on two problems in connexion with graphs.
\newblock \emph{Numerische mathematik}, 1\penalty0 (1):\penalty0 269--271,
  1959.

\bibitem[Drovandi and Pettitt(2013)]{drovandi2013bayesian}
Christopher~C Drovandi and Anthony~N Pettitt.
\newblock Bayesian experimental design for models with intractable likelihoods.
\newblock \emph{Biometrics}, 69\penalty0 (4):\penalty0 937--948, 2013.

\bibitem[Eriksson et~al.(2019)Eriksson, Pearce, Gardner, Turner, and
  Poloczek]{eriksson2019scalable}
David Eriksson, Michael Pearce, Jacob~R Gardner, Ryan Turner, and Matthias
  Poloczek.
\newblock Scalable global optimization via local bayesian optimization.
\newblock \emph{arXiv preprint arXiv:1910.01739}, 2019.

\bibitem[Foster et~al.(2019)Foster, Jankowiak, Bingham, Horsfall, Teh,
  Rainforth, and Goodman]{Foster2019-fj}
Adam Foster, Martin Jankowiak, Eli Bingham, Paul Horsfall, Yee~Whye Teh, Tom
  Rainforth, and Noah Goodman.
\newblock Variational bayesian optimal experimental design.
\newblock March 2019.

\bibitem[Frazier(2018)]{Frazier2018-bc}
Peter~I Frazier.
\newblock A tutorial on bayesian optimization.
\newblock July 2018.

\bibitem[Garnett et~al.(2010)Garnett, Osborne, and
  Roberts]{garnett2010bayesian}
Roman Garnett, Michael~A Osborne, and Stephen~J Roberts.
\newblock Bayesian optimization for sensor set selection.
\newblock In \emph{Proceedings of the 9th ACM/IEEE international conference on
  information processing in sensor networks}, pages 209--219, 2010.

\bibitem[Garnett et~al.(2012)Garnett, Krishnamurthy, Xiong, Schneider, and
  Mann]{garnett2012bayesian}
Roman Garnett, Yamuna Krishnamurthy, Xuehan Xiong, Jeff Schneider, and Richard
  Mann.
\newblock Bayesian optimal active search and surveying.
\newblock \emph{arXiv preprint arXiv:1206.6406}, 2012.

\bibitem[Hennig and Schuler(2012)]{Hennig2012-zu}
Philipp Hennig and Christian~J Schuler.
\newblock Entropy search for {Information-Efficient} global optimization.
\newblock \emph{J. Mach. Learn. Res.}, 13\penalty0 (57):\penalty0 1809--1837,
  2012.

\bibitem[Hern{\'a}ndez-Lobato et~al.(2014)Hern{\'a}ndez-Lobato, Hoffman, and
  Ghahramani]{Hernandez-Lobato2014-ec}
Jos{\'e}~Miguel Hern{\'a}ndez-Lobato, Matthew~W Hoffman, and Zoubin Ghahramani.
\newblock Predictive entropy search for efficient global optimization of
  black-box functions.
\newblock June 2014.

\bibitem[Houlsby et~al.(2012)Houlsby, Huszar, Ghahramani, and
  Hern{\'a}ndez-lobato]{Houlsby2012-xf}
Neil Houlsby, Ferenc Huszar, Zoubin Ghahramani, and Jose~M
  Hern{\'a}ndez-lobato.
\newblock Collaborative gaussian processes for preference learning.
\newblock In F~Pereira, C~J~C Burges, L~Bottou, and K~Q Weinberger, editors,
  \emph{Advances in Neural Information Processing Systems 25}, pages
  2096--2104. Curran Associates, Inc., 2012.

\bibitem[Kandasamy et~al.(2019)Kandasamy, Neiswanger, Zhang, Krishnamurthy,
  Schneider, and Poczos]{Kandasamy2019-qz}
Kirthevasan Kandasamy, Willie Neiswanger, Reed Zhang, Akshay Krishnamurthy,
  Jeff Schneider, and Barnabas Poczos.
\newblock Myopic posterior sampling for adaptive goal oriented design of
  experiments.
\newblock volume~97 of \emph{Proceedings of Machine Learning Research}, pages
  3222--3232, Long Beach, California, USA, 2019. PMLR.

\bibitem[Kleinegesse and Gutmann(2019)]{Kleinegesse2019-xg}
Steven Kleinegesse and Michael~U Gutmann.
\newblock Efficient bayesian experimental design for implicit models.
\newblock In Kamalika Chaudhuri and Masashi Sugiyama, editors,
  \emph{Proceedings of the Twenty Second International Conference on Artificial
  Intelligence and Statistics}, volume~89 of \emph{Proceedings of Machine
  Learning Research}, pages 476--485. PMLR, 2019.

\bibitem[Kleinegesse et~al.(2020)Kleinegesse, Drovandi, and
  Gutmann]{Kleinegesse2020-rh}
Steven Kleinegesse, Christopher Drovandi, and Michael~U Gutmann.
\newblock Sequential bayesian experimental design for implicit models via
  mutual information.
\newblock March 2020.

\bibitem[Krause et~al.(2008)Krause, Singh, and Guestrin]{Krause2008-zu}
Andreas Krause, Ajit Singh, and Carlos Guestrin.
\newblock {Near-Optimal} sensor placements in gaussian processes: Theory,
  efficient algorithms and empirical studies.
\newblock \emph{J. Mach. Learn. Res.}, 9\penalty0 (8):\penalty0 235--284, 2008.

\bibitem[Letham et~al.(2020)Letham, Calandra, Rai, and Bakshy]{letham2020re}
Ben Letham, Roberto Calandra, Akshara Rai, and Eytan Bakshy.
\newblock Re-examining linear embeddings for high-dimensional bayesian
  optimization.
\newblock \emph{Advances in Neural Information Processing Systems}, 33, 2020.

\bibitem[Li et~al.(2005)Li, Cheng, Hadjieleftheriou, Kollios, and
  Teng]{li2005trip}
Feifei Li, Dihan Cheng, Marios Hadjieleftheriou, George Kollios, and Shang-Hua
  Teng.
\newblock On trip planning queries in spatial databases.
\newblock In \emph{International symposium on spatial and temporal databases},
  pages 273--290. Springer, 2005.

\bibitem[Lindley(1956)]{lindley1956measure}
Dennis~V Lindley.
\newblock On a measure of the information provided by an experiment.
\newblock \emph{The Annals of Mathematical Statistics}, pages 986--1005, 1956.

\bibitem[Ma et~al.(2014)Ma, Garnett, and Schneider]{ma2014active}
Yifei Ma, Roman Garnett, and Jeff Schneider.
\newblock Active area search via bayesian quadrature.
\newblock In \emph{Artificial intelligence and statistics}, pages 595--603.
  PMLR, 2014.

\bibitem[Ma et~al.(2015)Ma, Sutherland, Garnett, and Schneider]{ma2015active}
Yifei Ma, Dougal Sutherland, Roman Garnett, and Jeff Schneider.
\newblock Active pointillistic pattern search.
\newblock In \emph{Artificial Intelligence and Statistics}, pages 672--680.
  PMLR, 2015.

\bibitem[Madsen(1973)]{madsen1973root}
Kaj Madsen.
\newblock A root-finding algorithm based on newton's method.
\newblock \emph{BIT Numerical Mathematics}, 13\penalty0 (1):\penalty0 71--75,
  1973.

\bibitem[Matthews et~al.(2017)Matthews, Van Der~Wilk, Nickson, Fujii,
  Boukouvalas, Le{\'o}n-Villagr{\'a}, Ghahramani, and
  Hensman]{matthews2017gpflow}
Alexander G de~G Matthews, Mark Van Der~Wilk, Tom Nickson, Keisuke Fujii,
  Alexis Boukouvalas, Pablo Le{\'o}n-Villagr{\'a}, Zoubin Ghahramani, and James
  Hensman.
\newblock Gpflow: A gaussian process library using tensorflow.
\newblock \emph{J. Mach. Learn. Res.}, 18\penalty0 (40):\penalty0 1--6, 2017.

\bibitem[M{\"u}ller(2005)]{Muller2005-lh}
Peter M{\"u}ller.
\newblock Simulation based optimal design.
\newblock In D~K Dey and C~R Rao, editors, \emph{Handbook of Statistics},
  volume~25, pages 509--518. Elsevier, January 2005.

\bibitem[Neiswanger et~al.(2014)Neiswanger, Wang, Ho, and
  Xing]{neiswanger2014modeling}
Willie Neiswanger, Chong Wang, Qirong Ho, and Eric~P Xing.
\newblock Modeling citation networks using latent random offsets.
\newblock In \emph{Proceedings of the Thirtieth Conference on Uncertainty in
  Artificial Intelligence}, pages 633--642, 2014.

\bibitem[Nelder and Mead(1965)]{nelder1965simplex}
John~A Nelder and Roger Mead.
\newblock A simplex method for function minimization.
\newblock \emph{The computer journal}, 7\penalty0 (4):\penalty0 308--313, 1965.

\bibitem[Osborne et~al.(2012)Osborne, Garnett, Ghahramani, Duvenaud, Roberts,
  and Rasmussen]{osborne2012active}
Michael Osborne, Roman Garnett, Zoubin Ghahramani, David~K Duvenaud, Stephen~J
  Roberts, and Carl~E Rasmussen.
\newblock Active learning of model evidence using bayesian quadrature.
\newblock In \emph{Advances in neural information processing systems}, pages
  46--54, 2012.

\bibitem[Pandey et~al.(2019)Pandey, Bhanodia, Khamparia, and
  Pandey]{pandey2019comprehensive}
Babita Pandey, Praveen~Kumar Bhanodia, Aditya Khamparia, and Devendra~Kumar
  Pandey.
\newblock A comprehensive survey of edge prediction in social networks:
  Techniques, parameters and challenges.
\newblock \emph{Expert Systems with Applications}, 124:\penalty0 164--181,
  2019.

\bibitem[Pleiss et~al.(2018)Pleiss, Gardner, Weinberger, and
  Wilson]{pleiss2018constant}
Geoff Pleiss, Jacob Gardner, Kilian Weinberger, and Andrew~Gordon Wilson.
\newblock Constant-time predictive distributions for gaussian processes.
\newblock In \emph{International Conference on Machine Learning}, pages
  4114--4123. PMLR, 2018.

\bibitem[Pleiss et~al.(2020)Pleiss, Jankowiak, Eriksson, Damle, and
  Gardner]{pleiss2020fast}
Geoff Pleiss, Martin Jankowiak, David Eriksson, Anil Damle, and Jacob~R.
  Gardner.
\newblock Fast matrix square roots with applications to gaussian processes and
  bayesian optimization.
\newblock In Hugo Larochelle, Marc'Aurelio Ranzato, Raia Hadsell,
  Maria{-}Florina Balcan, and Hsuan{-}Tien Lin, editors, \emph{Advances in
  Neural Information Processing Systems 33: Annual Conference on Neural
  Information Processing Systems 2020, NeurIPS 2020, December 6-12, 2020,
  virtual}, 2020.

\bibitem[Powell(1994)]{powell1994direct}
Michael~JD Powell.
\newblock A direct search optimization method that models the objective and
  constraint functions by linear interpolation.
\newblock In \emph{Advances in optimization and numerical analysis}, pages
  51--67. Springer, 1994.

\bibitem[Richardson(1911)]{richardson1911ix}
Lewis~Fry Richardson.
\newblock Ix. the approximate arithmetical solution by finite differences of
  physical problems involving differential equations, with an application to
  the stresses in a masonry dam.
\newblock \emph{Philosophical Transactions of the Royal Society of London.
  Series A, Containing Papers of a Mathematical or Physical Character},
  210\penalty0 (459-470):\penalty0 307--357, 1911.

\bibitem[Rios and Sahinidis(2013)]{rios2013derivative}
Luis~Miguel Rios and Nikolaos~V Sahinidis.
\newblock Derivative-free optimization: a review of algorithms and comparison
  of software implementations.
\newblock \emph{Journal of Global Optimization}, 56\penalty0 (3):\penalty0
  1247--1293, 2013.

\bibitem[Rosenbrock(1960)]{rosenbrock1960automatic}
HoHo Rosenbrock.
\newblock An automatic method for finding the greatest or least value of a
  function.
\newblock \emph{The Computer Journal}, 3\penalty0 (3):\penalty0 175--184, 1960.

\bibitem[Rubin(1984)]{rubin1984bayesianly}
Donald~B Rubin.
\newblock Bayesianly justifiable and relevant frequency calculations for the
  applies statistician.
\newblock \emph{The Annals of Statistics}, pages 1151--1172, 1984.

\bibitem[Seeger and Nickisch(2008)]{seeger2008large}
Matthias~W Seeger and Hannes Nickisch.
\newblock Large scale variational inference and experimental design for sparse
  generalized linear models.
\newblock \emph{arXiv preprint arXiv:0810.0901}, 2008.

\bibitem[Shahriari et~al.(2015)Shahriari, Swersky, Wang, Adams, and
  De~Freitas]{shahriari2015taking}
Bobak Shahriari, Kevin Swersky, Ziyu Wang, Ryan~P Adams, and Nando De~Freitas.
\newblock Taking the human out of the loop: A review of bayesian optimization.
\newblock \emph{Proceedings of the IEEE}, 104\penalty0 (1):\penalty0 148--175,
  2015.

\bibitem[Spall et~al.(1992)]{spall1992multivariate}
James~C Spall et~al.
\newblock Multivariate stochastic approximation using a simultaneous
  perturbation gradient approximation.
\newblock \emph{IEEE transactions on automatic control}, 37\penalty0
  (3):\penalty0 332--341, 1992.

\bibitem[Tran et~al.(2021)Tran, Neiswanger, Broderick, Xing, Schneider, and
  Ulissi]{tran2021computational}
Kevin Tran, Willie Neiswanger, Kirby Broderick, Eric Xing, Jeff Schneider, and
  Zachary~W Ulissi.
\newblock Computational catalyst discovery: Active classification through
  myopic multiscale sampling.
\newblock \emph{The Journal of Chemical Physics}, 154\penalty0 (12):\penalty0
  124118, 2021.

\bibitem[Villemonteix et~al.(2009)Villemonteix, Vazquez, and
  Walter]{villemonteix2009informational}
Julien Villemonteix, Emmanuel Vazquez, and Eric Walter.
\newblock An informational approach to the global optimization of
  expensive-to-evaluate functions.
\newblock \emph{Journal of Global Optimization}, 44\penalty0 (4):\penalty0
  509--534, 2009.

\bibitem[Wang and Jegelka(2017)]{Wang2017-fb}
Zi~Wang and Stefanie Jegelka.
\newblock Max-value entropy search for efficient bayesian optimization.
\newblock March 2017.

\bibitem[Williams and Rasmussen(2006)]{williams2006gaussian}
Christopher~KI Williams and Carl~E Rasmussen.
\newblock \emph{Gaussian processes for machine learning}, volume~2.
\newblock MIT press Cambridge, MA, 2006.

\bibitem[Wilson et~al.(2020)Wilson, Borovitskiy, Terenin, Mostowsky, and
  Deisenroth]{wilson2020efficiently}
James Wilson, Viacheslav Borovitskiy, Alexander Terenin, Peter Mostowsky, and
  Marc Deisenroth.
\newblock Efficiently sampling functions from gaussian process posteriors.
\newblock In \emph{International Conference on Machine Learning}, pages
  10292--10302. PMLR, 2020.

\bibitem[Zhong et~al.(2020)Zhong, Tran, Min, Wang, Wang, Dinh, De~Luna, Yu,
  Rasouli, Brodersen, et~al.]{zhong2020accelerated}
Miao Zhong, Kevin Tran, Yimeng Min, Chuanhao Wang, Ziyun Wang, Cao-Thang Dinh,
  Phil De~Luna, Zongqian Yu, Armin~Sedighian Rasouli, Peter Brodersen, et~al.
\newblock Accelerated discovery of co 2 electrocatalysts using active machine
  learning.
\newblock \emph{Nature}, 581\penalty0 (7807):\penalty0 178--183, 2020.

\end{thebibliography}





\newpage

\appendix

\section{}
\label{sec:appendix}

In this appendix, we give additional details about the acquisition functions
($\EIG_t^e(x)$ (\ref{eq:acqf_execpath}), $\EIG_t(x)$ (\ref{eq:acqf_output_expect}),
and $\EIG_t^v(x)$ (\ref{eq:acqf_subexecpath})),
discuss the computational cost of \infobaxnospace, and provide
additional experimental details.

\subsection{EIG for the Execution Path}
\label{sec:eigexepathappendix}

\paragraph{Details on equation (\ref{eq:postpred_given_exepath})}
Here, we justify more formally the statement given in
Eq.~(\ref{eq:postpred_given_exepath}), used to compute $\EIG_t^e(x)$,
that for an execution path sample 
$\exepathsamp \sim p(\exepath \given \datasett)$, where 
$\exepathsamp = (\samp{z}_s, \samp{f}_{z_s})_{s=1}^S$, then
\begin{align}
p \left(y_x \given \datasett, \exepathsamp \right)
    = p \left( y_x \; \Big| \; \datasett, \big\{ \samp{f}_{z_s} \big\}_{s=1}^S \right).
    \nonumber
\end{align}

The posterior predictive distribution conditioned on a posterior execution path
sample $\exepathsamp$ can be written
\begin{align}
p \left(y_x \given \datasett, \exepathsamp \right)
    &= p \left(y_x \given \datasett, \samp{z}_1, \samp{f}_{z_1}, \samp{z}_2, \samp{f}_{z_2},
        \ldots, \samp{z}_S, \samp{f}_{z_S} \right) \nonumber \\
    &= p \left(y_x \given \datasett, \samp{z}_1, \samp{f}_{z_1}, \samp{z}_2(\samp{z}_1, \samp{f}_{z_1}),
        \samp{f}_{z_2(\samp{z}_1, \samp{f}_{z_1})}, \ldots
        \right) \nonumber \\
    &= p \left(y_x \given \datasett, \samp{f}_{z_1},
        \samp{f}_{z_2(\samp{z}_1, \samp{f}_{z_1})}, \ldots 
        \right) 
\end{align}
where the third equality holds because each $\samp{z}_s = \samp{z}_s(\samp{z}_1, \samp{f}_{z_1}, \ldots)$
is a deterministic function of previous function evaluations
$\samp{f}_{z_1}, \ldots, \samp{f}_{z_{s-1}}$ in the sequence, as well as
the initial $\samp{z}_1$ (which is assumed to be a determinstic quantity
specified by algorithm $\Acal$),
so each $\samp{z}_s$ can be dropped from the conditioning.
Note also that the final line can be written equivalently as 
$p \left(y_x \given \datasett, \samp{f}_{z_1}, \samp{f}_{z_2}, \ldots \samp{f}_{z_S} \right)$
$=$
$p \left( y_x \; \Big| \; \datasett, \big\{ \samp{f}_{z_s} \big\}_{s=1}^S \right)$.


\paragraph{Background on Gaussian processes}
Gaussian processes (GPs) are popular models that are commonly used in Bayesian
optimization.
In order to give details on Eq.~(\ref{eq:postpred_given_exepath})
for a Gaussian process model, we first give background on GPs here.

A GP over the input space $\mathcal{X}$ is a random process characterized by a mean function
$\mu: \mathcal{X} \rightarrow \mathbb{R}$ and a covariance function (i.e. kernel)
$\kappa: \mathcal{X}^2 \rightarrow \mathbb{R}$.
If $f \sim \text{GP}(\mu, \kappa)$, then for all $x \in \mathcal{X}$, we can write
the distribution over $f$ at $x$ as  $f_x \sim \mathcal{N}(\mu(x), \kappa(x,x))$.
Suppose that we are given a dataset of $t$ observations $\datasett = \{(x_i, y_{x_i})\}_{i=1}^t$,
where
\begin{align}
    y_{x_i} = f_{x_i} + \epsilon_i \in \mathbb{R}
    \hspace{2mm} \text{and} \hspace{2mm}
    \epsilon_i \sim \mathcal{N}(0, \sigma^2).
\end{align}
Then the posterior process given $\datasett$ is also a GP, with mean function $\mu_t$ and 
covariance function $\kappa_t$, which we describe as follows.
Let $Y$, $k$, $k'  \in \Rbb^t$ be vectors where $Y_i = y_{x_i}$, $k_i = \kappa(x, x_i)$,
and $k'_i = \kappa(x', x_i)$.
Let $I_t \in \mathbb{R}^{t \times t}$ be the identity matrix
and let $K \in \mathbb{R}^{t \times t}$ be the \textit{Gram matrix}
with $K_{i,j} = \kappa(x_i, x_j)$. Then
\begin{align}
    \label{eq:gpposterior}
    \mu_t(x) & = k^\top (K + \sigma^2 I_t)^{-1} Y, \\
    \kappa_t(x, x') & = \kappa(x, x') - k^\top (K + \sigma^2 I_t)^{-1} k'.
\end{align}


Given $\datasett$, the posterior predictive distribution for a given $x \in \Xcal$, is 
$p(y_x \given \datasett) = \mathcal{N}(y_x \given \mu_x, \sigma_x^2)$, where
\begin{align}
    \label{eq:gppostpred}
    \mu_x = \mu_t(x)
    \hspace{2mm} \text{and} \hspace{2mm}
    \sigma_x^2 = \kappa_t(x, x) + \sigma^2.
\end{align}
For additional background on GPs, see \citet{williams2006gaussian}.

\paragraph{Equation (\ref{eq:postpred_given_exepath}) for Gaussian processes}
Under a GP model, we can derive a closed-form expression for Eq.~(\ref{eq:postpred_given_exepath}),
given  dataset $\datasett = \{(x_i, y_{x_i})\}_{i=1}^t$, and execution path sample
$\exepathsamp = (\samp{z}_s, \samp{f}_{z_s})_{s=1}^S$.
Intuitively, Eq.~(\ref{eq:postpred_given_exepath}) is the posterior predictive distribution for a GP
with two types of observations: noisy observations $y_{x_i}$ and noiseless observations $f_{z_s}$.
This can be written as
\begin{align}
    p \left(y_x \given \datasett, \exepathsamp \right) = 
    p \left( y_x \; \Big| \; \datasett, \big\{ \samp{f}_{z_s} \big\}_{s=1}^S \right) =
    \mathcal{N}\left(y_x \given \samp{\mu}_x, \samp{\sigma}_x^2 \right),
\end{align}
where we describe the two parameters $\samp{\mu}_x$ and $\samp{\sigma}_x^2 $ as follows.
Let $u = t + S$.
Let $\samp{Y} \in \Rbb^u$ be a vector where
\begin{align}
\samp{Y}_i =
    \begin{cases}
        y_{x_i},            & \text{if} \hspace{2mm} i \in \{1, \ldots, t \}\\
        \samp{f}_{z_{i-t}}  & \text{if} \hspace{2mm} i \in  \{t+1, \ldots, u\}.
    \end{cases}
\end{align}
Let $\samp{k} \in \Rbb^u$ be a vector where
\begin{align}
\samp{k}_i =
    \begin{cases}
        \kappa(x, x_i),                 & \text{if} \hspace{2mm} i \in \{1, \ldots, t \}\\
        \kappa(x, \samp{f}_{z_{i-t}})   & \text{if} \hspace{2mm} i \in  \{t+1, \ldots, u\},
    \end{cases}
\end{align}
and define $\samp{k}'$ similarly.
Let $I(\sigma) \in \mathbb{R}^{u \times u}$ be a diagonal matrix, where
\begin{align}
    I(\sigma)_{i,i} =
    \begin{cases}
        \sigma,     & \text{if} \hspace{2mm} i \in \{1, \ldots, t \}\\
        0           & \text{if} \hspace{2mm} i \in  \{t+1, \ldots, u\}.
    \end{cases}
\end{align}
Let $\samp{K} \in \mathbb{R}^{u \times u}$ be an extended \textit{Gram matrix}, where
\begin{align}
    \samp{K}_{i,j} =
    \begin{cases}
        \kappa(x_i, x_j),           & \text{if} \hspace{2mm} i, j \in \{1, \ldots, t\}\\
        \kappa(x_i, \samp{z}_{j-t}),       & \text{if} \hspace{2mm} i \in \{1, \ldots, t\},
            j \in \{t+1, \ldots, u\}\\
        \kappa(\samp{z}_{i - t}, x_j),     & \text{if} \hspace{2mm} i \in \{t+1, \ldots, u\},
            j \in \{1, \ldots, t\}\\
        \kappa(\samp{z}_{i-t}, \samp{z}_{j-t}),   & \text{if} \hspace{2mm} i, j \in \{t+1, \ldots, u\}.
    \end{cases}
\end{align}

Then
\begin{align}
    \samp{\mu}_x &= \samp{k}^\top (\samp{K} + I(\sigma))^{-1} \samp{Y}, \\
    \samp{\sigma}_x^2 &= \kappa(x, x) - \samp{k}^\top (\samp{K} + I(\sigma))^{-1} \samp{k}' + \sigma^2.
\end{align}

\subsection{EIG for the Algorithm Output}
\label{sec:eigoutputappendix}

In Eq.~(\ref{eq:acqf_output_expect}), the $\acqfnt(x)$ acquisition function is written
\begin{align}
    \acqfnt(x) =
    \entropyf{ y_x \given \datasett}
    - \expecf{p( \algoutput | \datasett)}
    {
        \entropyf{\expecf{p(\exepath | \algoutput, \datasett)}{
        p \left( y_x \given \datasett, \exepath \right)}}
    }. \nonumber
\end{align}
Here we describe details on how we estimate this acquisition function under a GP model.
In Section~\ref{sec:eigoutput}, we describe the general procedure:
we first draw a set of sample pairs
$P^j := \{(\exepathsamp^{\;j\;}, \algoutputsamp^{\;j\;})\}$,
consisting of an execution path and algorithm output,
by running algorithm $\Acal$
on samples $\samp{f} \sim p(f \given \datasett)$.
For a given output sample $\algoutputsamp^{\;j\;}$, we
then carry out an approximate Bayesian computation (ABC)-like
procedure to produce a set of execution path samples
\begin{align}
    \exepathring^{\;j\;} = \{ e \in P^j : \algoutputsamp \in \algoutputring^{\;j\;} \}, \nonumber
\end{align}
where $\algoutputring^{\;j\;}$ is a set of similar outputs defined as 
\begin{align}
    \algoutputring^{\;j\;} = 
    \left\{
    \outputplainsamp \in \{ \algoutputsamp^{\;k\;} \}_{k=1}^\ell
    \;:\;
    d (\outputplainsamp, \algoutputsamp^{\;k\;} ) < \delta, 
    \hspace{1mm} k \neq j
    \right\}, \nonumber
\end{align}
and where $d(\cdot, \cdot)$ is some distance function defined on the algorithm
output space $\Ocal$.
Note that, for a given $e \in \exepathring^{\;j\;}$, we can compute
$p(y_x \given \datasett, e )$ in 
closed form as described in Section~\ref{sec:eigexepathappendix}.
We can therefore estimate 
$\expecf{p(\exepath | \algoutput, \datasett)}{
        p \left( y_x \given \datasett, \exepath \right)}$
as a mixture density 
$\frac{1}{|\exepathring^{\;j\;}|} \sum_{e \in \exepathring^{\;j\;}} p(y_x \given \datasett, e ) $,
which in the case of GPs, 
will be a uniformly weighted mixture of Gaussians.
We can easily draw a set of $H$ samples from this mixture of Gaussians
to produce a set of one-dimensional samples
$\{\samp{y}_{x, 1}^{\;j\;}, \ldots, \samp{y}_{x, H}^{\;j\;}\} \subset \Rbb$,
and then construct a Monte Carlo estimate of the entropy via
$- \frac{1}{H} \sum_{h=1}^H \log \left(
\frac{1}{|\exepathring^{\;j\;}|}
\sum_{e \in \exepathring^{\;j\;}} p(\samp{y}_{x, h}^{\;j\;} \given \datasett, e )
\right)$.

By following these steps, we produce an estimate of
$\entropyf{\expecf{p(\exepath | \algoutputsamp^{\;j\;}, \datasett)}{
        p \left( y_x \given \datasett, \exepath \right)}}$ 
for $\algoutputsamp^{\;j\;} \sim p(\algoutput \given \datasett)$,
and then can follow the same procedure outlined in Section~\ref{sec:eigexepath} to
estimate the full $\acqfnt(x)$.


\subsection{EIG using an Execution Path Subsequence}
\label{sec:eigsubexeappendix}
Here, we give details on the acquisition function $\acqfnt^v(x)$ in Eq.~(\ref{eq:acqf_subexecpath}),
which is based on using a set of function values $\subexepath$ from a subsequence of the execution path.
To summarize, let the execution path $\exepath = (z_s, f_{z_s})_{s=1}^S$
have a subsequence of length $R$, denoted
$\subseqexepath := \subseqexepath(f) := (z_{i_r}, f_{z_{i_r}})_{r=1}^R$.
We can denote the \textit{function values} for this subsequence with
$\subexepath := \subexepath(f)$ $:=$ $\{f_{z_{i_r}}\}_{r=1}^R$.

We focus on the special case where the algorithm output $\algoutput$
exactly specifies this subsequence $\subseqexepath$, as well as its
function values $\subexepath$ (i.e.  $\subseqexepath$ and $\subexepath$ are
both a deterministic function of $\algoutput$, and are not random conditioned
on $\algoutput$).
There are a number of common applications where we can find such a subsequence, such as 
in optimization (where $\subseqexepath$ consists of an optima $x^*$ and its value $f_{x^*}$),
level set estimation (where $\subseqexepath$ is the set of $(x, f_x)$ pairs in a super/sublevel set),
root finding (where $\subseqexepath$ is a root of $f$), and phase mapping
(where $\subseqexepath$ is a set of $(x, f_x)$ pairs that comprise a phase boundary).
Additionally, the three applications that we show in
Section~\ref{sec:experiments}---estimating shortest paths in graphs, Bayesian local
optimization, and top-$k$ estimation---also fall into this setting.
In the first case, the subsequence is the sequence of edges and edge-costs that
comprise the minimum-cost path in a graph, and in the latter two cases, the
subsequence is the optima and associated function value(s).

Given this subsequence $\subseqexepath$, and its corresponding function values
$\subexepath$, we then propose using the following acquisition function:
\begin{align}
    \acqfnt^v(x)
    &= \hspace{2mm} \entropyf{y_x \given \datasett }
    - \expecf{p( f | \datasett)}
    { \entropyf{ y_x \given \datasett, \subexepath} }. \nonumber \\
    &= \hspace{2mm} \entropyf{y_x \given \datasett }
    - \expecf{p( f | \datasett)}
    { \entropyf{ y_x \given \datasett, \{f_{z_{i_r}}\}_{r=1}^R } }. \nonumber
\end{align}

Under a GP model, we can compute this acquisition function in closed form, using 
Eq.~(\ref{eq:postpred_given_exepath}), originally derived for the 
$\acqfnt^e(x)$ acquisition function (note that this fact is 
not true if we want to compute the $\EIG$ with respect to
the subsequence $\subseqexepath$, and that, in general, the posterior predictive
$p \left( y_x \given \datasett, \subseqexepath \right)$
$\neq$ $p \left( y_x \given \datasett, \subexepath \right)$).

Next we discuss why this acquisition function
shows strong performance in practice.
Ideally, we would like to determine $\subexepath$ such that
$\acqfnt^v(x)$ best approximates $\acqfnt(x)$. We can see that
\begin{align}
    \acqfnt^v(x) - \acqfnt(x) =
    \expecf{p(f \given \Dcal_t)}{
        \entropyf{y_x \given \Dcal_t, \subexepath}
        - \entropyf{y_x \given \Dcal_t, \algoutput}
    }.
\end{align}
So it is sufficient for us to determine a $\subexepath(f)$ such that 
$| \entropy{y_x \given \Dcal_t, \subexepath(\samp{f})}$ $-$
$\entropy{y_x \given \Dcal_t, \algoutput(\samp{f})} |$ is small
for all $\samp{f}$. Note also that
\begin{align}
    p(y_x \given \Dcal_t, \subexepathsamp) &=
        \expecf{p(\algoutput | \subexepathsamp, \Dcal_t)}{
            p(y_x \given \Dcal_t, \subexepathsamp, \algoutput)
        },
        \hspace{1mm} \text{and} \\
    p(y_x \given \Dcal_t, \algoutputsamp) &=
        \expecf{p(\subexepath | \algoutputsamp, \Dcal_t)}{
            p(y_x \given \Dcal_t, \subexepath, \algoutputsamp)
        }.
\end{align}

Intuitively, we would like a $\subexepath(f)$ such that
$\entropyf{\subexepath \given \algoutput, \Dcal_t}$
and $\entropyf{\algoutput \given \subexepath, \Dcal_t}$ are both zero, in which case
$p(y_x \given \Dcal_t, \subexepathsamp)$ $=$
$p(y_x \given \Dcal_t, \subexepathsamp, \algoutputsamp)$ $=$
$p(y_x \given \Dcal_t, \algoutputsamp)$,
and therefore $\acqfnt^v(x) = \acqfnt(x)$.
Interestingly, in our special case setting,
for a given sample $\samp{f} \sim p(f | \datasett)$ with
associated $\algoutputsamp = \algoutput(\samp{f})$ and
$\subexepathsamp = \subexepath(\samp{f})$, we find that
\begin{align}
    p(y_x \given \datasett, \algoutputsamp)
    = \expecf{p(\subexepath | \algoutputsamp, \datasett)}{
            p(y_x \given \datasett, \subexepathsamp, \algoutput)
        }
    = p(y_x \given \datasett, \subexepathsamp, \algoutputsamp) \nonumber
\end{align}
because the sample $\algoutputsamp$ deterministically specifies the 
execution path subsequence $\subexepathsamp$.
Thus, $\acqfnt^v(x)$ will serve as a good approximation to $\acqfnt(x)$
when  $\entropyf{\algoutput \given \subexepath, \datasett}$ is small,
and will be optimal when
$\entropyf{\algoutput \given \subexepath, \datasett} = 0$,
in which case $\acqfnt^v(x) = \acqfnt(x)$.

\subsection{Computational Considerations}
\label{sec:computational}
Our sampling-based approximation of the EIG objectives from Eqs.~(\ref{eq:acqf_execpath}),
(\ref{eq:acqf_output_expect}), and (\ref{eq:acqf_subexecpath}) require drawing posterior
samples from \(p(f_x | \datasett)\) and \(p(f_x | \datasett, \exepath)\). For Gaussian
processes, posterior sampling takes cubic time in the length of vector we condition on.
Thus this cost can be prohibitive for algorithms with long execution paths.

However, in our experiments, we rely on an implementation of GPU-accelerated, approximate
posterior sampling by the authors of \cite{wilson2020efficiently}, which can be found at 
\url{https://github.com/j-wilson/GPflowSampling}, which reduces the sampling complexity
to being linear in the length of the vector we condition on. Alternate methods for drawing
fast approximate GP posterior samples include \cite{pleiss2018constant} and \cite{pleiss2020fast}.
In our implementation, on a NVIDIA 1080ti GPU, drawing all samples from \(p(f_x \given \datasett)\)
and \(p(f_x \given \datasett, \exepath)\) takes only a few seconds at most, for each iteration of
\infobaxnospace, even when the execution path \(\exepath\) exceeds 8000 points, as in the case
of Dijkstra's algorithm.

\begin{figure*}
    \centering
    \includegraphics[height=1.7in]{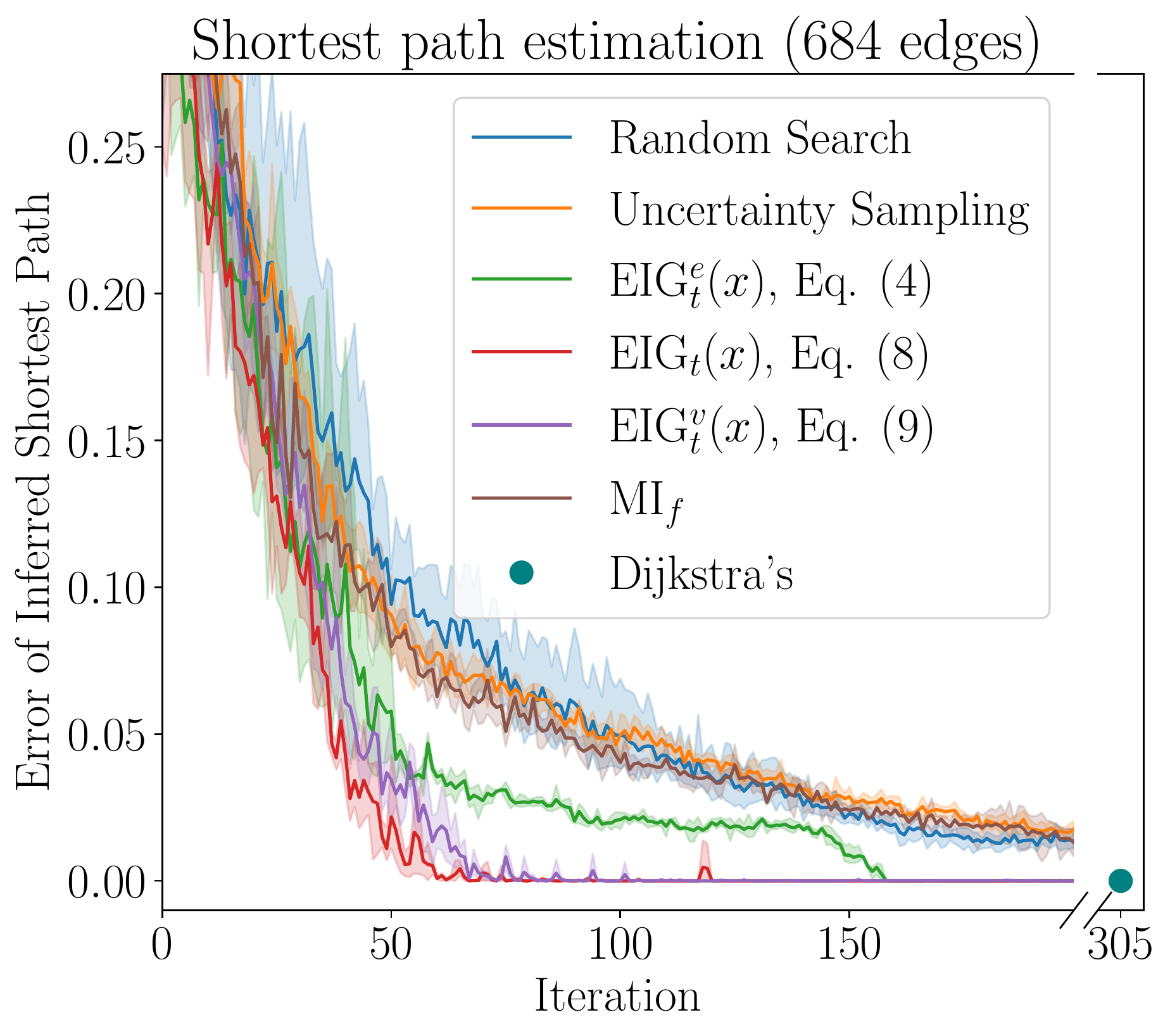}
    \hspace{-5pt}
    \includegraphics[height=1.7in]{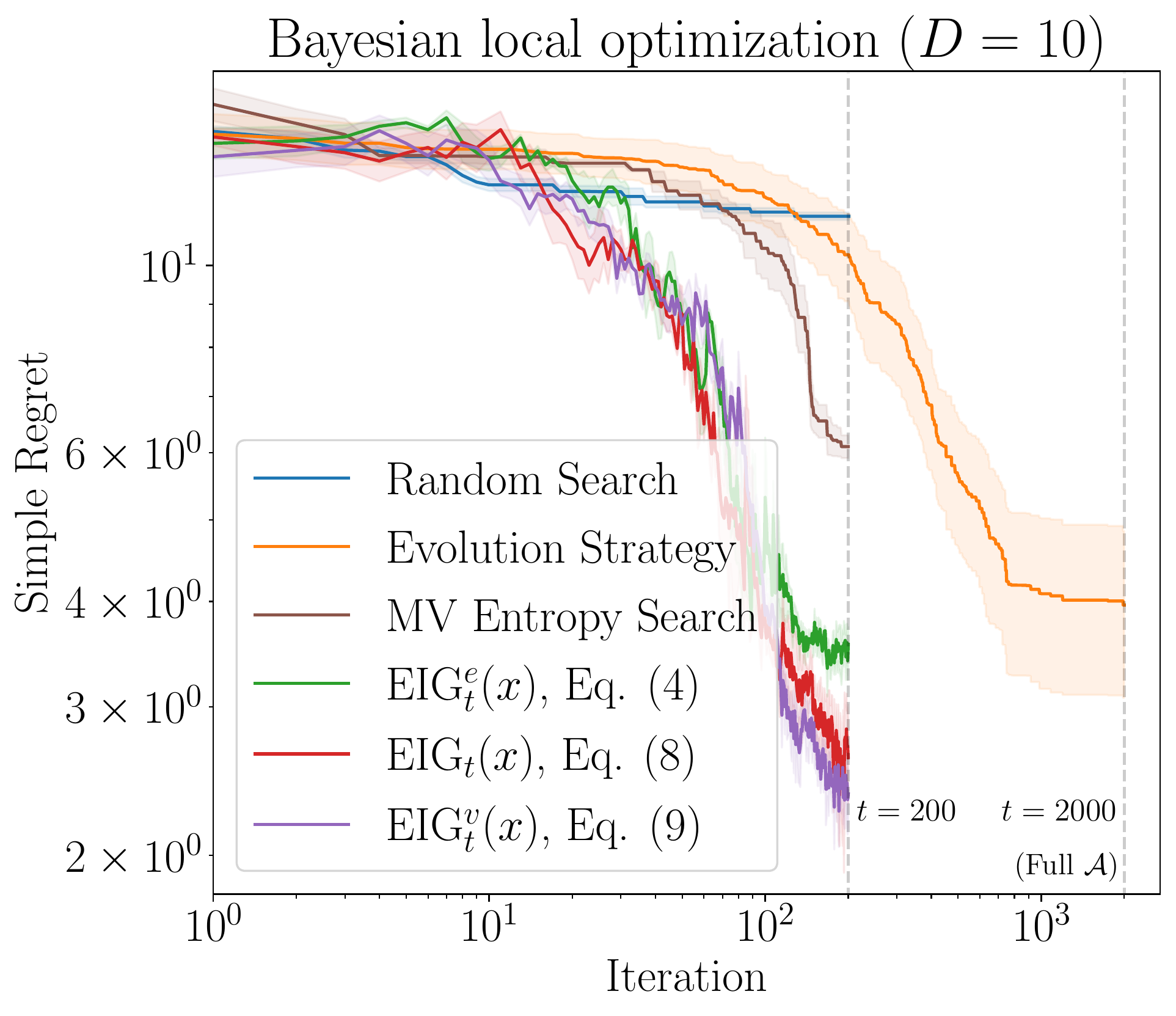}
    \hspace{-3pt}
    \includegraphics[height=1.7in]{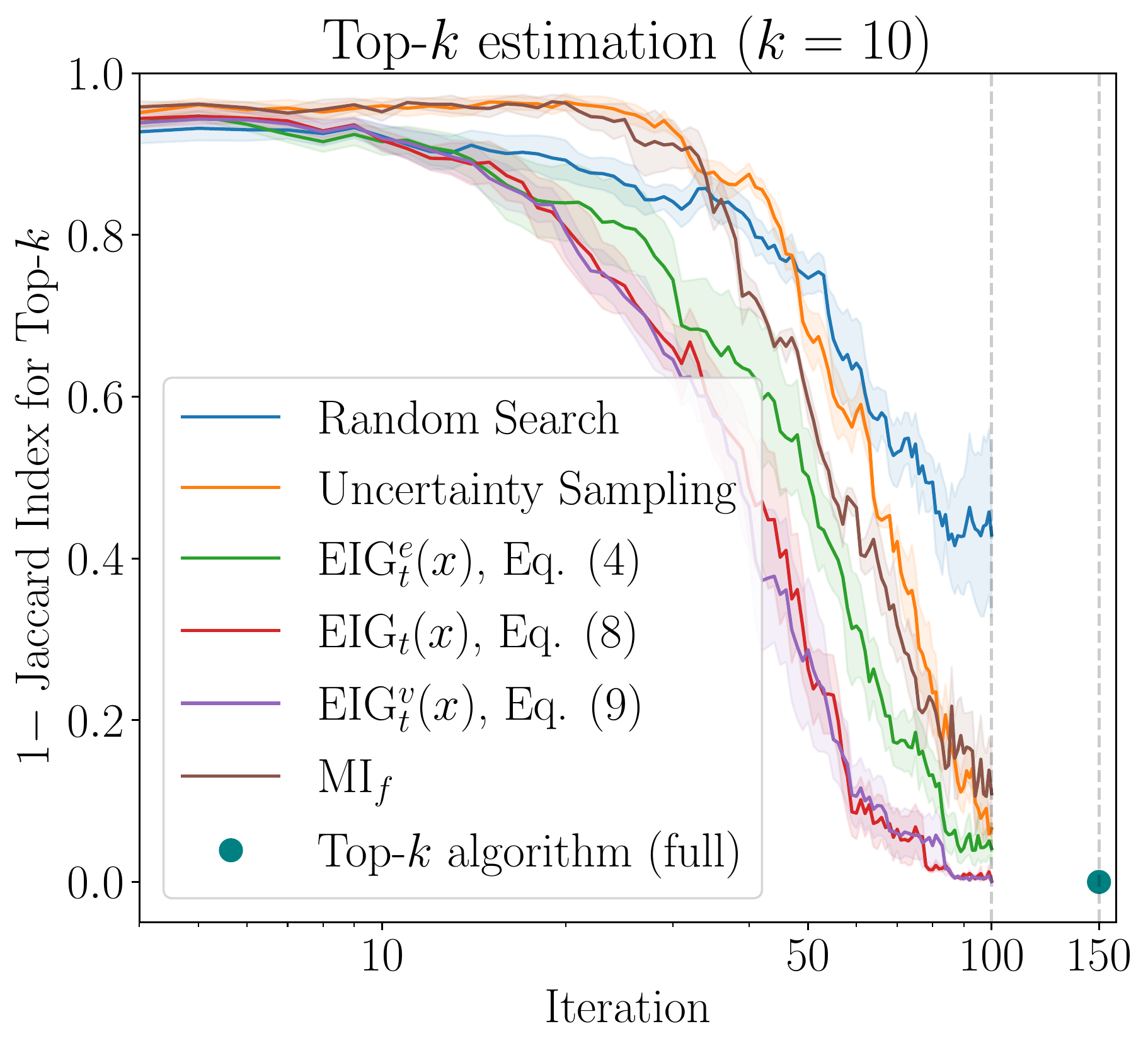}
    \caption{
    \textbf{Comparison of acquisition functions:} Comparison of \infobax performance using
    the three proposed acquisition functions $\acqfnt^e$ (\ref{eq:acqf_execpath}),
    $\acqfnt$ (\ref{eq:acqf_output_expect}), and $\acqfnt^v$ (\ref{eq:acqf_subexecpath}),
    along with baseline methods, on the applications of shortest path estimation,
    Bayesian local optimization, and top-$k$ estimation.
    }
    \label{fig:additionalresults}
    \vspace{-4mm}
\end{figure*}

\subsection{Additional Experimental Details}
\label{sec:experimentsappendix}

Here we include experimental details for the applications given 
in Section~\ref{sec:experiments}, and show additional experimental results, including on a 
comparison of proposed acquisition functions.
Note that we use a Gaussian processes as our prior \(p(f)\) for all experiments.

\paragraph{Comparison of acquisition functions}
In Figure~\ref{fig:additionalresults} we show the results of experiments where we compare the
three estimators we proposed in Eqs.~(\ref{eq:acqf_execpath}), (\ref{eq:acqf_output_expect}),
and (\ref{eq:acqf_subexecpath}), on shortest path estimation, Bayesian local optimization, and top-$k$ estimation.
Each plot shows a measure of error on the $y$-axis, and number of iterations (i.e. queries) on the $x$-axis.
In all three cases, we see that \infobax using $\acqfnt^v(x)$ and $\acqfnt(x)$ tend to perform best,
followed closely by \infobax using $\acqfnt^e(x)$, and afterwards by the baseline methods.
In the shortest path and top-$k$ estimation plots, we have also included an additional baseline, denoted
$\text{MI}_f$, which sequentially chooses queries that maximize the expected information gain about the 
function $f$ (which has similarities with the \textsc{UncertaintySampling} baseline).

\paragraph{Details on estimating shortest paths}
Here, we give details about our first experimental application, 
on estimating shortest paths in graphs (Section~\ref{sec:experiments}).
Note that some examples of real-world networks with potentially expensive
edge query lookups include transportation networks
(e.g. querying or negotiating the price of transport, or assessing the cost of travel)
\citep{bast2016route},
social networks (e.g. measuring the amount of social connection or targeted similar
interests) \citep{backstrom2011supervised, pandey2019comprehensive},
and article networks (e.g. assessing the relevance of papers based on content)
\citep{adafre2005discovering, neiswanger2014modeling}.

For the grid-shaped graph, to define edge costs, we use a rescaled Rosenbrock function 
\citep{rosenbrock1960automatic} for $\Xcal \subset \Rbb^2$, defined as
\begin{align}
    f(x) &= 10^{-2}[((a-x_2)^2 + b(x_2 - x_1^2)^2)]
\end{align}
with \(a = 1, b=100\) within the domain \(-2 \leq x_1 \leq 2\), \(-1 \leq x_2 \leq 4\). Each vertex within the grid is connected to its closest neighbors, corresponding to the ordinal and cardinal directions as shown in Figure \ref{fig:dijkstra_area}.

We create our California graph from a subset of the graph network provided by \cite{li2005trip}, corresponding to the region within \(36.7^\circ\)N and \(39.3^\circ \)N and \(122.5^\circ\)W and \(119.5^\circ\)W. We use the vertex at \(36.1494^\circ\)N, \(122.045158^\circ\)W as our start vertex and the vertex at \(38.913666^\circ\)N, \(120^\circ\)W as our destination vertex. These two positions correspond to roughly Santa Cruz and Lake Tahoe, respectively. To obtain the ground truth edge cost function, we take the average of the cooridinates of the two vertices at each end of the edge and query its elevation from the OpenElevation dataset. To ensure that the edge costs are non-negative, we first rescale all edge costs by the max edge cost and add an offset of $0.1$ to each edge cost. 

To ensure that our distribution $p(f)$ is supported on only non-negative functions, we transform the edge costs through the inverse of the softplus function and fit our Gaussian process on these transformed edge costs which can take on negative values.
In all cases, when running \infobax, we draw $20$ posterior samples of the shortest path. We found that drawing more samples did not speed up convergence to the true shortest path.

We compute the $\acqfnt^v(x)$ acquisition function in Eq.~(\ref{eq:acqf_subexecpath}),
with respect to an execution path subsequence.
In this case, the algorithm output is a sequence of edges and their respective edge costs,
where each edge is associated with a point in \(\Xcal\),
i.e. the average position between the two vertex positions.
We therefore use the costs of the edges along a shortest path sample output
as a $\subexepathsamp$ in this acquisition function.

To evaluate the quality of each inferred shortest path, we use the 2D polygonal area between the inferred path and the truth shortest path. To do this, we decompose the area into a set of disjoint 2D polygons, and compute the area of each polygon using the shoelace algorithm (i.e. Gauss's area formula).

\paragraph{Details on Bayesian local optimization}
Here, we give details about our second experimental application,
on Bayesian local optimization (Section~\ref{sec:experiments}).
In this application, we demonstrate the use of \infobax for the task
of black-box optimization, where the algorithm $\Acal$ is a local optimization algorithm---i.e.
an algorithm consisting (typically) of an iterative procedure that returns some local optima
with respect to a given initialization.
Intuitively, the goal is to perform black-box optimization by choosing a sequence of
function evaluations which efficiently yield a good estimate of the output of $\Acal$ 
(rather than, for example, choosing evaluations to directly infer a global optima of the function).

For our local optimization algorithm $\Acal$, we use a mutation-based evolution strategy.
In this algorithm, 
we first initialize a population of $p$ vectors $V^p = \{v_j\}_{j=1}^p$ (where $v_j \in \Xcal$)
all to the same point, which is drawn uniformly at random from $\Xcal$.
The algorithm then proceeds over a sequence of $g$ generations.
At each generation, we mutate each vector in this population via a normal
proposal, i.e. draw $\widetilde{v}_j \sim \mathcal{N}(v_j, \sigma_{pr}^2)$
and set $v_j \leftarrow \widetilde{v}_j$, for all $v_j \in V^p$.
We then query the function $f(v_j)$, for each 
$v_j \in V^p$, and discard the bottom $(1 - e)\%$ (where $e \in [0, 1]$)
of vectors in $V^p$ based on their function values,
before proceeding on to the next generation.
%
After $g$ generations, we return the vector $v_j^*$ that achieved the
best queried function value over the course of the full algorithm
(i.e. over all generations),
and it's observed function value $f(v_j^*)$.
We refer to this algorithm as \textsc{EvolutionStrategy}.

We show experimental results on minimization of three standard benchmark functions:
Branin ($\Xcal \subset \Rbb^2$),
Hartmann-6 ($\Xcal \subset \Rbb^6$),
and Ackley-10 ($\Xcal \subset \Rbb^{10}$), defined as
\begin{align}
    \text{Branin:} \hspace{4mm}
    f(x) &= \left( x_2 - \frac{5.1}{(4 \pi)^2}x_1^2 + \frac{5}{\pi}x_1 - 6 \right)^2
            + 10 \left( 1-\frac{1}{8 \pi} \right) \cos(x_1) + 10
    \nonumber \\
    \text{Hartmann-6:} \hspace{4mm}
    f(x) &= - \sum_{i=1}^4 \alpha_i \exp \left( -\sum_{j=1}^6 A_{ij} (x_j - B_{ij})^2 \right)
    \nonumber \\
    \text{Ackley-10:} \hspace{4mm}
    f(x) &=
    - 20 \exp \left( -\frac{1}{5} \sqrt{\frac{1}{10} \sum_{i=1}^{10} x_i^2} \right)
    - \exp \left( \frac{1}{10} \sum_{i=1}^{10} \cos(2 \pi x_i) \right)
    + 20 + \exp(1),
    \nonumber
\end{align}
where, in Hartmann-6,
\begin{align}
    \alpha &= (1.0, 1.2, 3.0, 3.2)^\top\\
    A &= \left(
    \begin{matrix}
    10      & 3     & 17    & 3.5   & 1.7   & 8\\
    0.05    & 10    & 17    & 0.1   & 8     & 14\\
    3       & 3.5   & 1.7   & 10    & 17    & 8\\
    17      & 8     & 0.05  & 10    & 0.1   & 14 
    \end{matrix}
    \right)
    \nonumber\\
    B &= \left(
    \begin{matrix}
    1312    & 1696  & 5569  & 124   & 8283  & 5886\\
    2329    & 4135  & 8307  & 3736  & 1004  & 9991\\
    2348    & 1451  & 3522  & 2883  & 3047  & 6650\\
    4047    & 8828  & 8732  & 5743  & 1091  & 381
    \end{matrix}
    \right) \times 10^{-4}.
    \nonumber
\end{align}

We compare the performance of four methods: \infobaxnospace, \textsc{RandomSearch}
(described in Section~\ref{sec:experiments}), \textsc{EvolutionStategy},
and \textsc{MaxValueEntropySearch} \citep{Wang2017-fb}, which is a popular
information-based Bayesian optimization method that aims to efficiently
infer a global optima of $f$.

In \infobaxnospace, we draw 100 samples of $\samp{f} \sim p(f \given \datasett)$ 
and run $\Acal$ on these, in order to produce the 
execution path samples $\exepathsamp$.
We then use the $\acqfnt^v(x)$ acquisition function 
from Eq.~(\ref{eq:acqf_subexecpath}). Since the algorithm output 
consists of a vector $v_j^* \in \Xcal$, and its value $f(v_j^*)$,
we use this tuple, $(v_j^*, f(v_j^*))$, as the execution path 
subsequence in $\acqfnt^v(x)$. 
For acquisition optimization in both \infobax and
\textsc{MaxValueEntropySearch}, we run a high-iteration random search algorithm.
In \textsc{MaxValueEntropySearch}, for our global
optimization procedure, we also run a high-iteration random search algorithm.

As an error metric, for each method we compute the Simple Regret, defined as
the difference between the function value $f(\hat{x})$ at an estimated 
minimum $\hat{x}$ and the true minimal value $f(x^*)$. For each method,
we therefore need to produce an estimated minimum $\hat{x}$. For
\textsc{EvolutionStrategy}, the estimated minimum is chosen to be the output
of the algorithm (described above).
For \textsc{RandomSearch}, the estimated minimum is chosen to be the input with 
lowest queried value, i.e. $\hat{x} := \argmax_{x \in \datasett} f(x)$.
For both \infobax and \textsc{MaxValueEntropySearch}, the estimated
minimum is chosen to be the estimated local or global optimum (respectively) of the 
GP posterior mean $\expecf{p(f | \datasett)}{f}(x) = \mu_T(x)$.



\end{document}